\documentclass[12pt]{article}

 \usepackage{setspace}
\usepackage{fancyhdr}
\usepackage{subcaption}
\usepackage{lscape}
\usepackage[authoryear]{natbib}
\usepackage{algorithmic}
\usepackage{algorithm}
\usepackage{tikz}
\usepackage[margin=1in]{geometry}

\usepackage[colorlinks = citecolor]{hyperref}
\usepackage{booktabs}
\usepackage{amsmath}
\usepackage{amssymb}
\usepackage{mathtools}
\usepackage{amsthm}

\usepackage[capitalize,noabbrev]{cleveref}

\theoremstyle{plain}
\newtheorem{theorem}{Theorem}[section]

\newtheorem{lemma}[theorem]{Lemma}
\newtheorem{corollary}[theorem]{Corollary}
\theoremstyle{definition}
\newtheorem{definition}[theorem]{Definition}
\newtheorem{assumption}[theorem]{Assumption}
\theoremstyle{remark}

\begin{document}

\title{\vspace{-40pt} Unique Rashomon Sets for Robust Active Learning}

\author{Simon Nguyen\\University of Washington\\simondn@uw.edu  \and Kentaro Hoffman\\University of Washington\\khoffm3@uw.edu \and Tyler McCormick\\University of Washington\\tylermc@uw.edu}
\date{}

\maketitle

\begin{abstract}
    Collecting labeled data for machine learning models is often expensive and time-consuming. Active learning addresses this challenge by selectively labeling the most informative observations, but when initial labeled data is limited, it becomes difficult to distinguish genuinely informative points from those appearing uncertain primarily due to noise. Ensemble methods like random forests are a powerful approach to quantifying this uncertainty but do so by aggregating all models indiscriminately. This includes poor performing models and redundant models, a problem that worsens in the presence of noisy data. We introduce UNique Rashomon Ensembled Active Learning (\textit{UNREAL}), which selectively ensembles only distinct models from the Rashomon set, which is the set of nearly optimal models. Restricting ensemble membership to high-performing models with different explanations helps distinguish genuine uncertainty from noise-induced variation. We show that \textit{UNREAL} achieves faster theoretical convergence rates than traditional active learning approaches and demonstrates empirical improvements of up to 20\% in predictive accuracy across five benchmark datasets, while simultaneously enhancing model interpretability.
\end{abstract}

\section{Introduction}

Collecting labeled data to train data-hungry modern artificial intelligence (AI) and machine learning (ML) models can be expensive or time-consuming. This challenge arises in a wide range of applications: image labeling \citep{ZhuBento2017, HuijserGemert2017}, sentence classification \citep{ALMQS_Schumann2019}, and verbal autopsy \citep{setel2007scandal,mccormick2016probabilistic,jha2019automated,PPI_VerbalAutopsy}. In such scenarios, strategically determining which observations are most informative for model training will greatly reduce data redundancy and improve the model's ability to generalize from limited data.

To address time and budget constraints, active learning provides a framework where observations are adaptively and strategically selected for labeling based on their potential informativity. The key task in active learning is choosing the most informative observations that will enhance the predictive quality of the model when labeled. However, this selection process faces a fundamental challenge: with limited labeled data initially available, how can we confidently identify truly informative observations rather than noise? 
% This challenge requires a principled approach to uncertainty quantification, particularly in the presence of label noise.

Amongst the many metrics of informativity, \citet{Uncertainty_LewisGale1994, freund1997selective, ErrorReduction_Roy2001}, uncertainty is the most commonly employed \citep{LiuSuvey}. One particularly effective approach to quantifying uncertainty is through query-by-committee (QBC), where multiple models form a committee whose disagreement in predictions indicates which observations are most uncertain and therefore informative to label. Ensemble methods such as random forests are especially well-suited for active learning with QBC techniques \citep{freund1997selective, DiverseEnsembleAL} as their weak learners naturally form a diverse committee. This diversity is crucial as it incorporates different explanations of the data into the committee, with disagreement among ensemble members then serving as a direct measure of uncertainty and informativity \citep{VoteEntropyHard, QBCDiversity_Kee2018}.

While ensemble methods offer a natural way to quantify uncertainty through the diversity of multiple learners (models), ensemble methods tend to aggregate over the space of \textit{all} models, even if some of the models may have relatively poor accuracy. While some approaches such as Bayesian model averaging account for this by weighting the models by how likely they are given the data \citep{BMA}, having a large number of mediocre models can make such weighting approaches difficult, especially in cases of limited, high-dimensional, or noisy data \citep{GRAEFE2015943}. %Aggregating poor and implausible models compromises the query-selection criteria, potentially leading to a suboptimal query in the active learning process.

This challenge is further exacerbated in the presence of label noise, which although common in realistic settings, is well-known to be detrimental to active learning strategies \citep{FasterRatesOfConvergence_CNW}. Active learning strategies, designed to prioritize querying points where models are most uncertain, may overfit to noisy labels instead of focusing on true areas of uncertainty. This challenge makes incorporating a diverse set of explanations among committee members even more crucial.

To address these limitations, we propose an algorithm that enhances active learning by restricting aggregation to a subset of well-performing and high-evidence models known as the Rashomon set. Importantly, in the presence of label noise, the Rashomon set naturally expands to encompass a more diverse set of valid explanations, providing a principled way to capture and utilize the inherent ambiguity in noisy data. By ensembling the models within the Rashomon set, which is the set of nearly optimal models, our approach ensures that the active learning process is driven by models with high evidence while accounting for noise through multiple diverse yet equally valid explanations of the data.

Throughout the paper, we rely on tree-based models.  These models partition the feature space into groups with similar outcomes.  Tree-based models are widely used in practice and have been extensively explored, particularly in combination with a sparsity requirement, in the context of Rashomon sets.  However, a common and widely known issue with tree-based models is that multiple trees can represent the same partition.  

The multiplicity of trees creates two distinct challenges. First, it means that more complex partitions are more likely to appear in the Rashomon set simply by chance (because as the complexity of the partition increases, there are more possible trees that represent the same partition).  A similar issue occurs in discrete Bayesian model averaging with uniform priors over models--there are more medium sized models, so the ``flat'' prior actually strongly favors models of a particular size.  Second, the multiplicity of trees means that approaches that sample from the space of trees expend additional, unnecessary, computational cycles evaluating redundant trees.  This issue arises because trees enforce an ordering, or hierarchy among variables that have no natural order.  ~\cite{AparaRPS} address this issue by using a Hasse diagram, which, unlike trees, is a geometric object compatible with partially ordered sets.      
 
In this paper, we remain in the space of trees and demonstrate that one can further restrict the Rashomon set to a unique set of classification patterns while still preserving the performance of the active learning process. This Rashomon-based method, titled UNique Rashomon Ensembled Active Learning (\textit{UNREAL}), ensures that the ensemble incorporates the interpretability of weak learners by only including the set of unique and diverse explanations. Simulations on five benchmark datasets demonstrate improvements in prediction accuracy compared to standard approaches.

\section{Rashomon Sets}\label{RashomonSetSection}
When constructing machine learning models, researchers face two distinct types of uncertainty. The first originates from the variability in the predicted outcomes generated by a given model, often referred to as a model's \textit{intrinsic risk}. The second originates from selecting the right model from a vast and diverse hypothesis space, a phenomenon known as \textit{model ambiguity}. This distinction, originally articulated by economist Kenneth Arrow in 1951, separates the uncertainty of prediction from a given model from the uncertainty of choosing among many plausible models \citep{Arrow1951}. 

Many modern machine learning approaches have become extremely good at reducing intrinsic risk within a model, but often fail to fully account for model ambiguity. Methods such as LASSO search for a single optimal model and conclusions tend by drawn based on the sole optimal model. This ignores any potential for model ambiguity, as there may in fact exist many models with similar predictive power for a given dataset \citep{AmazingThingsGoodModels}. On the other hand, ensemble methods such as Bayesian Model Averaging \citep{BMA} aggregate models across the full space of models, but this involves soliciting input from many models that are implausible given the observed data \citep{10.1214/ss/1009212519}.  To choose either of these approaches, an analyst would need to have extreme preferences.  Choosing a single model indicates absolutely no aversion to ambiguity in the model, whereas averaging over all models (even those that look exceedingly unlikely given the data at hand) would indicate an extreme aversion to the possibility of selecting the wrong model.

The Rashomon Effect posits the existence of near-optimal models that have similarly high predictive performance, but arrive at predictions in different and diverse ways. This phenomenon exposes a core issue in the current machine learning paradigm: reliance on a single predictive model is overly sensitive to specific data patterns and conceals the potential existence of alternative equally valid explanations \citep{RashomonSimplerMLModels, AmazingThingsGoodModels}. %

To quantify the impact of Rashomon Effect, we introduce the Rashomon set.
\begin{definition}[Rashomon set]\label{def:RashomonSet}
    Let $S$ be a given data set, $\mathcal{F}$ the space of all possible models, and $\phi$ the loss function. For a threshold $\epsilon \ge 0$, the Rashomon set $\hat{R}_\epsilon(\mathcal{F})$ is given by
    \begin{equation}
        \hat{R}_{\epsilon}(\mathcal{F}):= \{f \in \mathcal{F}: L(f) \le \hat{L}(\hat{f}) + \epsilon\}
    \end{equation}
    % \begin{equation}
    %     \hat{R}(\mathcal{F}, \epsilon):= \{f \in \mathcal{F}: L(f) \le (1+\epsilon)\cdot\hat{L}(\hat{f})\}
    % \end{equation}
    such that $\hat{f}$ is the empirical risk minimizer of $S$ with respect to the loss function $\phi: \hat{f} \in \arg \min_{f \in \mathcal{F}}L(f)$.
\end{definition}
Intuitively, the Rashomon set is the set of all plausible models that come within an $\epsilon$ distance of the empirical risk minimizer. By enumerating the Rashomon set, researchers can explore the full range of plausible explanations supported by the data. This range of diverse explanations is particularly valuable in noisy settings, in which the uncertainty across different models of the Rashomon set can highlight regions of the data where different reasonable models are giving ambiguous results.

Importantly, near-optimal does not mean identical. Models that share similar loss do not necessarily share the same classification pattern. Even if all the models in the Rashomon set have similar performance, they can (and often do) differ in how they partition the covariate space. If the models are trees, this will mean that trees with similar losses still partition the feature space differently. Multiple distinct trees in the Rashomon set may be similar in accuracy but disagree in specific regions of the feature space (see Figure \ref{fig:TreeGeometry} of the Appendix). That disagreement is particularly meaningful in active learning, where many algorithms rely on querying candidate observations from regions where near-optimal models diverge..

Despite this potential to identify meaningful disagreement regions among near-optimal models, implementing this insight presents practical challenges. Traditional ensemble methods such as random forests rely on randomization through the sampling of features and observations to form and aggregate base learners. While this randomization creates less-correlated trees, this often incorporates implausible base models that perform relatively poorly, introducing spurious disagreements that can cause the active learner to chase noise rather than true uncertainty. In contrast, Rashomon sets allow for the targeted aggregation of only high-performing models, reducing the risk of incorporating poor models in the query-selection process. This means that disagreements across the Rashomon ensemble are more likely to reflect more genuine and meaningful uncertainty in the data rather than random instability or overfitting to noise.

In the space of decision trees, \citet{ExploringRashomonTrees} is the first to provide an algorithm that completely enumerates the Rashomon set for sparse decision trees. Their algorithm, \texttt{TreeFarms}, provides an exhaustive yet computationally feasible method to generate, store, and view the entire Rashomon set of decision trees. However, due to the inherent structure and geometry of the decision trees, many trees in this set offer redundant expressions of the same partition of the covariate space, which becomes increasingly problematic in active learning, as it has the potential to further skew our metric of uncertainty in the committee by artificially inflating agreement \citep{DiverseEnsembleAL}. This can be seen in Figure \ref{fig:ErrorbyTreeIndex_Grouped} and more deeply in Figure \ref{fig:TreeGeometry} of the Appendix. To address this limitation, Section 4 will outline our method of grouping trees based on their unique classification patterns.

\section{Active Learning}\label{ALSection}

\subsection{Notation}

Borrowing notation from \citet{LiuSuvey}, let observation $i$ be composed of data $(\mathbf{x}_i, y_i)$ for vector $\mathbf{x}_i$ in covariate space $\mathcal{X}$ and label $y_i$ in output space $\mathcal{Y}$. The data is sent through a supervised learning model $F(\cdot): \mathcal{X} \to \mathcal{Y}$. When $F(\cdot)$ is an ensemble method consisting of base learners, denote the base learners at $\{f_m\}_{m=1}^M$. The model is learned from a training dataset $D_{tr} = \{(\mathbf{x}_i,y_i)\}_{i=1}^I$ and tested by an independent dataset $D_{ts} = \{(\mathbf{x}_j,y_j)\}_{j=1}^J$. The goal is to train $F(\cdot)$ to predict the labels of the out-of-sample test set with a budget-constrained number of labeled observations.

Active learning seeks to adaptively and strategically choose which unlabeled observations should be queried for oracle labeling and then be used in the supervised learning model. Let the query iteration in the active learning framework be denoted by $n$. Denote the reservoir of unlabeled candidate observations as $D_{cdd}^{(n)}= \{(\mathbf{x}_k,y_k)\}_{k=1}^K$ with $y_k$ initially unknown. A selector $S(\cdot)$ is the strategy used to select samples from $D_{cdd}^{(n)}$ to be Oracle labeled. At each iteration, $S(\cdot)$ will sample one observation, denoted $B^{(n)}$, from the candidate dataset $D_{cdd}^{(n)}$ without replacement to query for oracle labeling. $B^{(n)}$ is then added to the training set and removed from the candidate set: $D_{cdd}^{(n+1)} = D_{cdd}^{(n)} \cup B^{(n)}$ and $D_{cdd}^{(n+1)} = D_{cdd}^{(n)}\backslash B^{(n)}$. The model is then retrained on the new training set as $F^{{(n+1)}}\big(D_{tr}^{(n+1)}\big)$. As such, the $B^{(n)}$ is chosen to find the observations that are most informative to improving predictive performance. 

The process is repeated, gradually expanding the training set with informative observations, until the labeling budget is reached or a desired classification metric threshold is met. 

\subsection{Query-By-Committee Metrics}
Picking a selector metric is a key topic in the active learning literature. Common methods are uncertainty \citep{Uncertainty_LewisGale1994}, query-by-committee metrics \citep{freund1997selective, Settles}, or expected error \citep{ErrorReduction_Roy2001}. Due to the ensembling nature of our methods, we choose to measure informativity by query-by-committee (QBC) metrics, particularly Argamomn-Engelson and Dagan's vote entropy \citep{VoteEntropyHard}:
\begin{equation}\label{eq:HardVoteEntropy}
    \delta(y,\mathbf{x},\mathcal{C}) = \max_\mathbf{x} - \sum_{y \in \mathcal{Y}} \frac{\text{vote}_{\mathcal{C}}(y,\mathbf{x})}{|\mathcal{C}|}\log \frac{\text{vote}_{\mathcal{C}}(y,\mathbf{x})}{|\mathcal{C}|}
\end{equation}
where $\text{vote}_{\mathcal{C}}(y,\mathbf{x}) = \sum_{c \in \mathcal{C}} \mathbb{I}\{c(\mathbf{x}) = y\}$ is the number of ``votes" that label $y \in \mathcal{Y}$ receives for $\mathbf{x}$ amongst the members $c$ of committee $\mathcal{C}$. 

This selector metric is a committee-based generalization of uncertainty measures that consider the confidence of each committee member. It can be viewed as a Bayesian adaptation of Shannon's 1948 uncertainty sampling entropy \citep{Shannon1948}. 
One can observe from Equation \ref{eq:HardVoteEntropy} that ensembling duplicate models has the potential to overinflate the vote entropy with trees from the best-performing explanation group \citep{DiverseEnsembleAL}. This weakness will be addressed by including diverse, yet unique, classification patterns in our proposed methodology \textit{UNREAL}.

\section{From Noise to Signal: The Rashomon Advantage}\label{NoiseSection}

Active learning strategies suffer greatly in the presence of label noise \citep{FasterRatesOfConvergence_CNW, Burbridge, DAL_Noise, Hetero_Noise, DIRECT_NOISE}. Specifically, label noise decreases the accuracy of each supervised learning model in the committee and thus increases the possibility of suboptimal models. This leads to a less optimal active learning strategy in general. While one option is to expand to the committee to include noise-resistant models, we employ a different strategy based on Rashomon theory which does not require finding and implementing such noise-resistant models.

\citet{RashomonSimplerMLModels} and \citet{RashomonRatioNoise} discovered that increasing the label noise while keeping the sample size fixed increases the number of models that exhibit similar performance, i.e. the size of the Rashomon set. Specifically, they argue that label noise leads to increased pattern diversity - the average difference in predictions between distinct classification patterns in the Rashomon set. The increase in pattern diversity manifests as predictive multiplicity \citep{PredictiveMultiplicityMarx}, indicating that there are more differences in model predictions and thus more diverse models in the Rashomon set. Thus, in the presence of noise, the distinction of being the ``best-fitting" model becomes increasingly arbitrary, as multiple models capture different, yet equally valid patterns in the data. 

This enhanced model diversity suggests that relying on a single ``best" model may overlook other meaningful explanations of the data. Active learning, in its iterative addition of training data, presents a unique opportunity to leverage this existence of a diverse yet equally valid model. Rashomon sets offer a natural defense against noise in active learning: instead of relying on a single model that may overfit to noise points, we can leverage the entire set of plausible explanations to capture a comprehensive view of uncertainty in the dataset. That is, incorporating multiple valid explanations from the Rashomon set can help distinguish between genuine uncertainty and noise-induced variability in the sampling strategy. 

To this end, \textit{UNREAL} proposes that one should query \textit{within} the Rashomon Set. Rather than expanding a committee's diversity through different model classes, \textit{UNREAL} enriches the committee by incorporating multiple models from within the same model class that exhibit similar predictive performance. Specifically, for a given model class, \textit{UNREAL} constructs a committee from the Rashomon set $\mathcal{\hat{R}}_\epsilon$, consisting of all models achieving accuracy within $\epsilon$ of the best-performing model. This approach ensures that each committee member represents a valid alternative explanation of the underlying data pattern, rather than merely capturing noise. Active learning then proceeds following the standard QBC framework with uncertainty quantified through vote entropy \citep{VoteEntropyHard}.

The empirical effectiveness of this approach comes from three factors. Firstly, the models models in the Rashomon set, while all near-optimal, are not identical. The regions where they disagree are precisely the ones that matter most in active learning, as they provide the strongest and most meaningful signal of genuine uncertainty across all plausible models. Secondly, the Rashomon set provides a natural defense against one of active learning's main weakness: noise. By ensembling across a set of plausible models, the selector committee has a more robust view of uncertainty than if it had incorporated potentially suboptimal models. Finally, as the committee ensembles these near-optimal yet diverse models, the algorithm prioritizes querying points with true uncertainty.

Ensembling across the Rashomon sets represents a fundamental shift in how we approach active learning under noise. By leveraging the natural diversity of near-optimal models captured in the Rashomon set, we have a method that inherently adapts to the presence of noise. Rather than treating noise as an obstacle that confuses the active learning selection criteria, UNREAL harnesses the insights from Rashomon theory to turn model ambiguity into an advantage, allowing us to distinguish between genuine uncertainty and noise-induced variability.

An interesting potential area of future would be proving theoretically that restricting ourselves to the Rashomon set of models in active learning leads to provable improvements in convergence rates of the active learning algorithm.  One potential strategy could be to build on work such as \citet{FasterRatesOfConvergence_CNW}, which highlights the importance of exploring areas of meaningful diversity or uncertainty in the set of possible models.  A proof strategy could proceed by showing that the Rashomon set, with an appropriate $\epsilon$, faithfully captures such plausible diversity while regularizing away heterogeneity that is likely spurious. 

\section{\textit{UNREAL}: Unique Rashomon Ensemble Active Learning}\label{AlgorithmSection}

In our proposed approach, the committee $\mathcal{C}$ in Equation \ref{eq:HardVoteEntropy} is constructed as the Rashomon set of decision trees, denoted by $\mathcal{\hat{R}}_\epsilon$. This Rashomon set $\mathcal{\hat{R}}_\epsilon$ consists of ``near-equal" decision trees whose objective function is within $\epsilon$ of the overall best model given the data. Since each near-equal model in the Rashomon set provides a different perspective on the data, diverse prediction patterns emerge in regions of genuine ambiguity. Our method exploits this diversity by ensembling across this set of distinct explanations to capture uncertainty arising from the variety of plausible models.

Although the trees in the Rashomon set share many of the same splits and appear globally correlated across the dataset, the regions where they exhibit disagreements are crucial to identifying true uncertainty. This property is particularly valuable in active learning: the trees agree on easily classified points (avoiding wasteful queries) but diverge precisely in areas where additional labels would be most informative. As new labels are acquired in these contentious regions, the near-optimal trees can collectively converge to more decisive predictions, making this approach particularly effective for query-by-committee strategies.

To enumerate the Rashomon set of decision trees, we use \citet{ExploringRashomonTrees}'s \texttt{TreeFarms} approach. \texttt{TreeFarms} exhaustively enumerates the Rashomon set of decision trees, allowing us to aggregate the best models in our ensemble method. However, unlike random forests, \texttt{TreeFarms} lacks the random sampling of features and data, making the models in \texttt{TreeFarms} correlated. This correlation in decision trees presents a significant challenge for query-by-committee approaches, as correlation amongst committee members may both artificially inflate agreement in the vote and complicate interpretability \citep{DiverseEnsembleAL}.

To address this issue, we reduce redundancy in \texttt{TreeFarms} by grouping trees based on their unique classification patterns and selecting a single representative tree from each of these groups to ensemble. This ensures that each chosen tree is meaningfully distinct while faithfully representing the Rashomon set. This prevents overestimating agreement in our vote entropy metric, ultimately leading to a valid query-by-committee voting selection method. 

\begin{figure}[!htp]
    \centering
    \includegraphics[width=0.9\columnwidth]{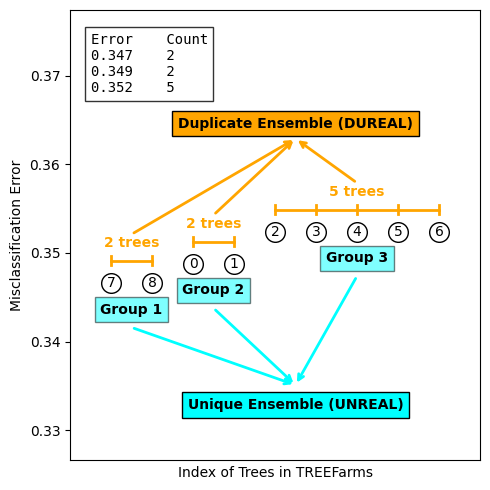}
    \caption{A depiction of the redundant and unique ensembling of Rashomon trees from the COMPAS dataset. To visualize our methods, the figure plots the misclassification errors by the ordered indices of the tree (though the unique selection of trees was grouped by classification pattern in our simulations). As shown, many trees have the \textit{same} misclassification rate, indicating these trees share the same classification pattern. The geometry of the trees can be seen in Figure \ref{fig:TreeGeometry} of the appendix.}
    \label{fig:ErrorbyTreeIndex_Grouped}
\end{figure}

Our approach can be visualized in Figure \ref{fig:ErrorbyTreeIndex_Grouped}. If we ignore the redundancy of explanations in \texttt{TreeFarms}, the ensemble will be dominated by the trees in Group 3 offering the same explanation and prediction. This will artificially inflate agreement amongst our committee (see Equation \ref{eq:HardVoteEntropy}) by overvaluing the trees of the most commonly occurring explanation groups. If we instead account for the redundancy of the trees, the unique selection method will instead choose one tree arbitrarily from each classification pattern, diversifying our committee and more fully representing the Rashomon set. Our method is summarized in Algorithm \ref{alg:UNREAL}.

\begin{algorithm}[!htb]
    \caption{Unique Rashomon Ensemble Active Learning} 
    \label{alg:UNREAL}
\begin{algorithmic}
   \STATE {\bfseries Input:} Training dataset $D_{tr}^{(0)}$; Test dataset $D_{ts}$; Candidate dataset $D_{cdd}^{(0)}$; Rashomon Threshold $\epsilon$;
   \REPEAT
        \STATE Enumerate and train the Rashomon set $\mathcal{\hat{R}}_\epsilon$ of models $\{f_m\}_{m=1}^M$ on $D^{(n)}_{tr}$ with \texttt{TreeFarms}.
        \STATE Generate the test set predicted labels $\hat{y}^{(n)}_{ts,m}$ for each model $f_m \in \mathcal{\hat{R}}_\epsilon$ from $\mathbf{x}_{ts}^{(n)}$.
        \STATE Ensemble the test set predicted labels: $\bar{\hat{y}}^{(n)}_{ts}:= \text{mode}\left(\hat{y}^{(n)}_{ts,m}\right)$.
        \STATE Evaluate the performance of the ensemble on $D^{(n)}_{ts}$ with prediction $\bar{\hat{y}}^{(n)}_{ts}$.
        \STATE Group models by identical classification patterns and select one representative from each group.
        \STATE Generate the candidate set predicted labels $\hat{y}^{(n)}_{cd,m}$ for each unique model $f_m \in \mathcal{\hat{R}}_\epsilon$ from $\mathbf{x}_{cd}^{(n)}$.
        \STATE Ensemble the candidate set predicted labels: $\bar{\hat{y}}^{(n)}_{cd}:= \text{mode}\left(\hat{y}^{(n)}_{cd,m}\right)$.
        \STATE Compute the vote-entropy metric $\delta^{(n)}(\bar{\hat{y}}^{(n)}_{cd},\mathbf{x}_{cd},\mathcal{\hat{R}}_\epsilon)$ from Equation \ref{eq:HardVoteEntropy}.
        \STATE Resample $B^{(n)}$ from $D_{cdd}^{(n)}$ based on the observation with the highest vote entropy:
        $B^{(n)} := \arg\max_{\mathbf{x}} \delta^{(n)}(\bar{\hat{y}}^{(n)}_{cd},\mathbf{x}_{cd},\mathcal{\hat{R}}_\epsilon)$
        \STATE Query $B^{(n)}$ for oracle labeling
        \STATE Set $D^{(n+1)}_{tr} = D^{(n)}_{tr} \cup B^{(n)}$ and $D_{cdd}^{(n+1)} = D_{cdd}^{(n)} \setminus B^{(n)}$   
   \UNTIL{labeling budget is depleted or test error is sufficiently small}
\end{algorithmic}
\end{algorithm}

\subsection{How to Choose the Rashomon Threshold}\label{ChoosingRashomonThreshold}

The Rashomon threshold $\epsilon$ provides a principled way to navigate the exploration-exploitation tradeoff inherent in active learning. A larger $\epsilon$ includes more diverse models in the committee, promoting exploration of alternative explanations, while a smaller $\epsilon$ restricts the committee to models closest to the empirical risk minimizer, favoring exploitation of established patterns. This choice significantly impacts how the algorithm handles uncertainty, particularly in the presence of noise.

The different choices for the size of the Rashomon set represent a trade-off between using the currently most accurate model versus the robustness of making decisions based on the wider range of near-optimal models. While our empirical experiments show that a good choice for the Rashomon model can strongly improve the active learning's accuracy and rate of convergence, this is predicated on being able to identify a good choice for the Rashomon threshold $\epsilon$.

The threshold effectively serves as a lever balancing between current accuracy and robustness. When $\epsilon$ is too small, the committee may consist of nearly identical models, oversampling regions that initially appear uncertain but may reflect only the limitations of the current best model. Conversely, when $\epsilon$ is too large, the committee may include relatively poor-performing models, potentially drowning genuine signals in a sea of noise.

\begin{figure}[!htp]
    \centering
    \includegraphics[width=0.5\columnwidth]{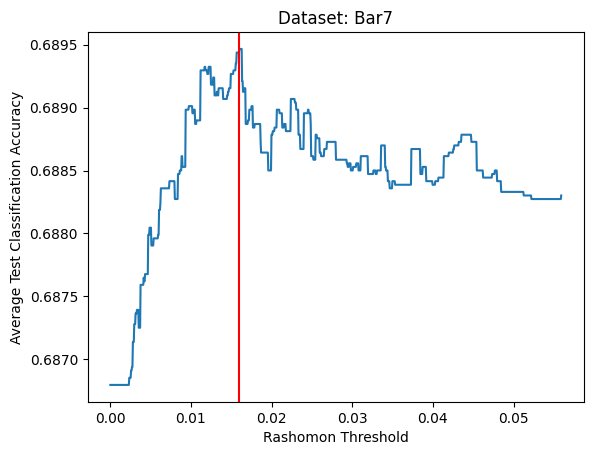}
    \caption{This plot graphs the ensemble test classification error against the Rashomon threshold. In this case, the optimal threshold is $0.016$.}
    \label{fig:OptimalRashomonThreshold_Bar7}
\end{figure}

Based on our empirical experiments, we recommend initializing the Rashomon threshold to a value that is \textit{at or larger than} the optimal Rashomon size for the initial training data. As an example, in Figure \ref{fig:OptimalRashomonThreshold_Bar7} we have the average accuracy over all the models in the Rashomon set as the size of the Rashomon set is varied. All trees were estimated using \texttt{TreeFarms} using a random $20\%$ training split of the data on the Bar7 dataset. We observe that as we increase the Rashomon threshold, and hence the size of the Rashomon set, the average classification accuracy on the test set increases up until $\epsilon = 0.016$ and then decays. As such, $\epsilon =0.016$ should be chosen as the value for our Rashomon threshold throughout the active learning procedure.

This procedure can be improved, albeit at the cost in terms of sample size or computational cost. One can split the training set in two, training the models on one, and computing accuracies to choose the Rashomon cutoff with the other. Such data splits in general are good practice to prevent overfitting \citep{ElementsStatisticalLearning}. However, in cases where labeled data is scarce, as is in the case of active learning, one may lack sufficient data to reserve for such splits. In such cases, we recommend one err on the side of caution and choose a Rashomon threshold that is slightly larger than what may be suggested by the initial dataset. 

\section{Experiments}\label{Experiments}
\subsection{Datasets}\label{DataSets}
Empirical experiments were performed on five benchmark datasets: Iris \citep{IrisDataset}, MONK-1, MONK-3 \citep{MONKDataset}, COMPAS dataset \citep{COMPASDataset}, and Bar7 \citep{Bar7Dataset}. The summary of the data as well as preprocessing details can be seen in Table \ref{tab:DatasetTable} of Section \ref{SimulationDetailsAppendixSection} in the appendix.

The datasets vary in complexity of their data-generating functions. MONK-1 and Iris represent relatively simple underlying structures. MONK-3 introduces a more intricate generative scheme but is still a fixed rule. The Bar7 and COMPAS datasets are characterized by the most notable noise and complexity due to their collection in restaurant and recidivism settings. These five datasets provide a gradient of learning challenges, enabling nuanced evaluation of our proposed methods against other active learning methods under varying degrees of complexity and noise.

\subsection{Simulation}
To compare our unique selection of classification patterns to baseline methods, we ran the active learning process against QBC with random forests and passive learning, in which observations are randomly selected for labeling. We also compared our method to a QBC method that ensembles \textit{all} of the trees of the Rashomon set, whose method we call \textit{DUplicate Rashomon Ensemble Active Learning (DUREAL)}. \textit{DUREAL} is essentially a weighted variant of \textit{UNREAL} in which classification patterns are weighted by how frequently they appear in the Rashomon set. Due to the intrinsic geometry of decision trees, trees with larger complexity (ie. more splits and depth) will appear more frequently in our committee.

One hundred active learning runs were run on the Iris, MONK-1, and MONK-3 datasets, fifty on COMPAS, and fifteen on Bar 7 due to computational limitations. The Rashomon threshold values used in \textit{UNREAL} and \textit{DUREAL} are given in Table \ref{tab:DatasetTable}. Each method had a fixed regularization of $0.01$ on the depth of the trees. Random forests were run with $100$ weak learners. Twenty percent of each dataset was reserved for the test set with twenty percent of the remaining observations used as the initial training dataset. We evaluate our active learning procedures with the F1 score to account for both precision and recall. Further information about the simulation can be seen in Section \ref{SimulationDetailsAppendixSection} of the appendix. Open source code for the simulation is available on \href{https://github.com/thatswhatsimonsaid/RashomonActiveLearning}{Github}.
\subsection{Experimental Results}

%%% FIVE BY ONE FIGURES %%%

\begin{figure}
    \centering
    %%% Left Column %%%
    \begin{minipage}[t]{0.45\textwidth}
        %%% Row 1: Iris  %%%
        \begin{subfigure}[t]{\columnwidth}
            \centering
            \includegraphics[width=\columnwidth]{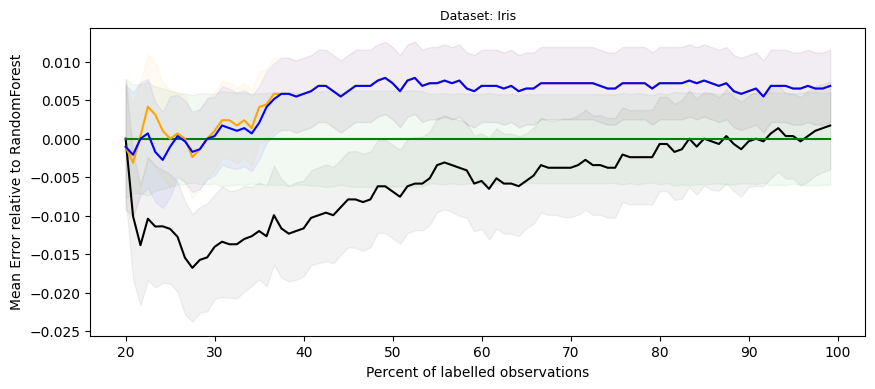}
        \end{subfigure}
        \vspace{1em}
        %%% Row 2: MONK1  %%%
        \begin{subfigure}[t]{\columnwidth}
            \centering
            \includegraphics[width=\columnwidth]{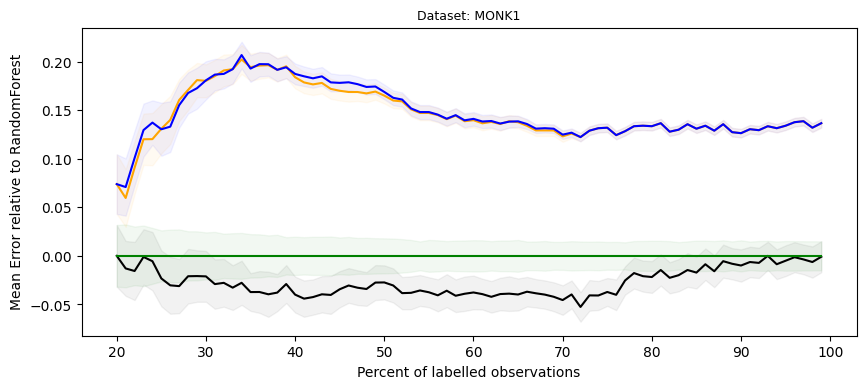}
        \end{subfigure}
        \vspace{1em}
        %%% Row 3: MONK3  %%%
        \begin{subfigure}[t]{\columnwidth}
            \centering
            \includegraphics[width=\columnwidth]{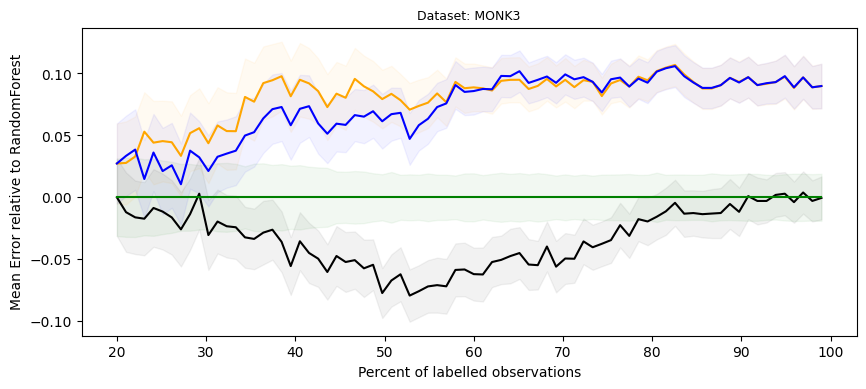}
        \end{subfigure}
    \end{minipage}
    \hfill
    %%% Right Column %%%
    \begin{minipage}[t]{0.45\textwidth}
        %%% Row 4: Bar7 %%%
        \begin{subfigure}[t]{\columnwidth}
            \centering
            \includegraphics[width=\columnwidth]{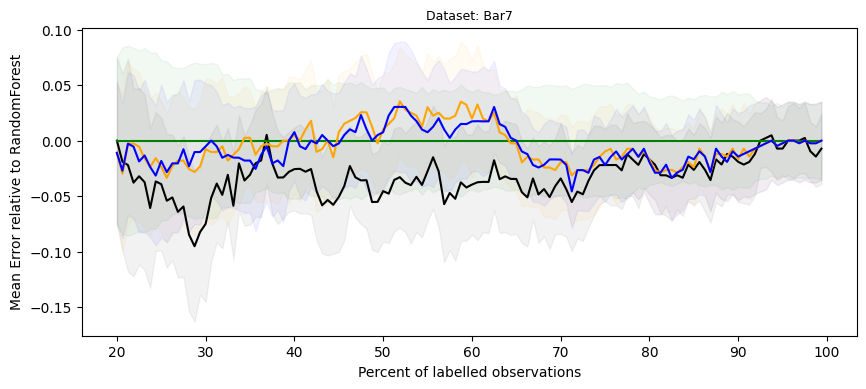}
        \end{subfigure}
        \vspace{1em}
        %%% Row 5: COMPAS  %%%
        \begin{subfigure}[t]{\columnwidth}
            \centering
            \includegraphics[width=\columnwidth]{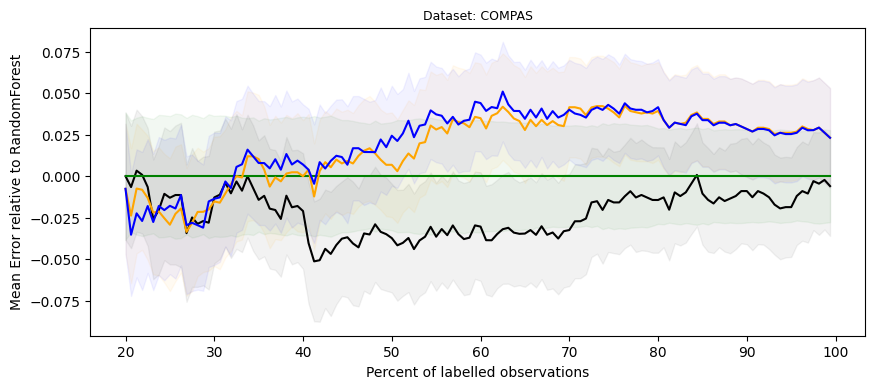}
        \end{subfigure}
        \vspace{3em} % Extra space to align with left column's 3 figures
    \end{minipage}
    
    %%% Legend below all plots %%%
    \begin{tikzpicture}
        % Define custom colors using RGB from Python
        \definecolor{PassiveColor}{rgb}{0, 0, 0}  % Black
        \definecolor{RFColor}{rgb}{0, 0.5, 0}     % Adjusted Green
        \definecolor{DUREALColor}{rgb}{1, 0.5, 0} % Orange
        \definecolor{UNREALColor}{rgb}{0, 0, 1}   % Blue
        
        % Adjusted positions for even spacing across figure width
        \draw[PassiveColor, thick] (0,0) -- (0.5,0) node[right, black] {Passive};
        \draw[RFColor, thick] (2.0,0) -- (2.5,0) node[right, RFColor] {RF};
        \draw[DUREALColor, thick] (3.5,0) -- (4,0) node[right, DUREALColor] {DUREAL};
        \draw[UNREALColor, thick] (6.0,0) -- (6.5,0) node[right, UNREALColor] {UNREAL};
    \end{tikzpicture}
    
    %%% Caption and Label %%%
    \caption{Performance of the four active learning procedures (left) on our five benchmark datasets. The plot presents the errors relative to random forests.}
    \label{fig:ActiveLearningErrorFigure}
\end{figure}
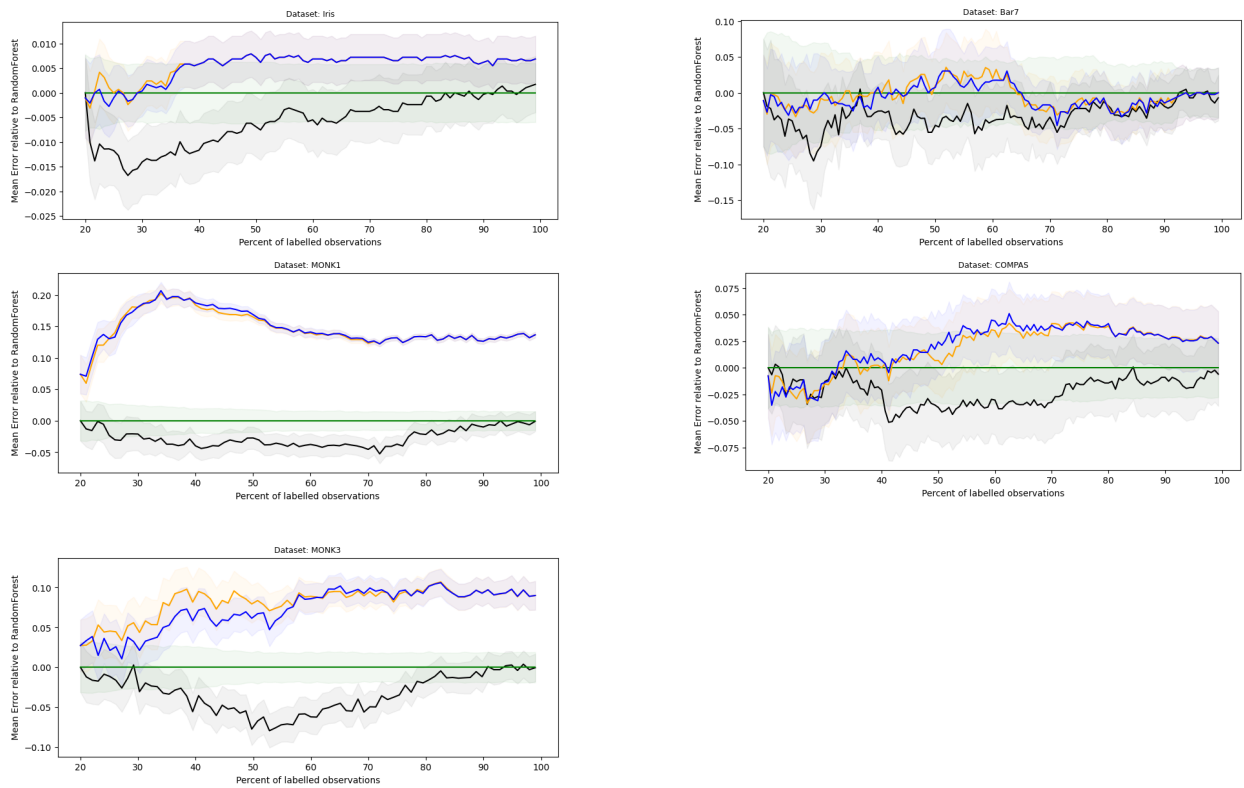

Results are presented in Figure \ref{fig:ActiveLearningErrorFigure}, showing errors relative to our random forest baseline. Standard error plots and Wilcoxon rank signed tests at 95\% significance level are provided in Figure \ref{fig:ActiveLearningErrorFigureRelativeToNone} and Table \ref{tab:WRST_All} of the appendix, respectively.

Our findings demonstrate the remarkable gains when ensembling across the Rashomon set and the substantial advantages over traditional QBC with random forests. Ensembles of decision trees from the Rashomon set (both unique and duplicate) consistently achieve superior performance, even up to $20\%$ in the MONK-1 dataset, highlighting the robustness and prediction accuracy achievable with this approach. With the exception of Bar7, we note that this gap becomes increasingly larger as more data is added to our training set.
% Notably, our methods exhibit stronger performance in the initial stages of the active learning procedure where training data is most limited. This "head start" can be attributed to our method's selective aggregation of only the highest-performing trees of the Rashomon set.

Interestingly, we find that at times \textit{DUREAL} mildly outperforms \textit{UNREAL}, contrary to concerns about artificial committee agreement inflation. This can be explained by the combinatorial nature of tree splits: classification patterns with greater complexity tend to appear more frequently. This creates an implicit weighted voting system in \textit{DUREAL} where complex trees carry greater influence. Nonetheless, \textit{UNREAL}'s compartmentalization of redundant explanations maintained similarly high classification accuracy while reducing to a more parsimonious ensemble of learners.

\begin{figure}
    \centering
    %%% Left Column %%%
    \begin{minipage}[t]{0.45\textwidth}
        %%% Row 1: Iris  %%%
        \begin{subfigure}[t]{\columnwidth}
            \centering
            \includegraphics[width=\columnwidth]{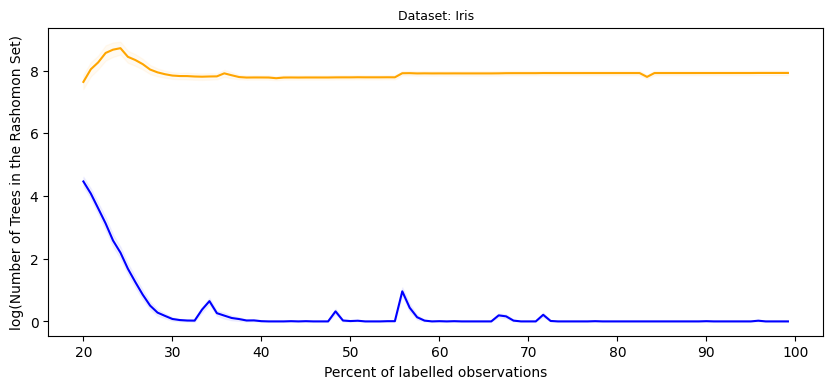}
        \end{subfigure}
        \vspace{1em}
        %%% Row 2: MONK1  %%%
        \begin{subfigure}[t]{\columnwidth}
            \centering
            \includegraphics[width=\columnwidth]{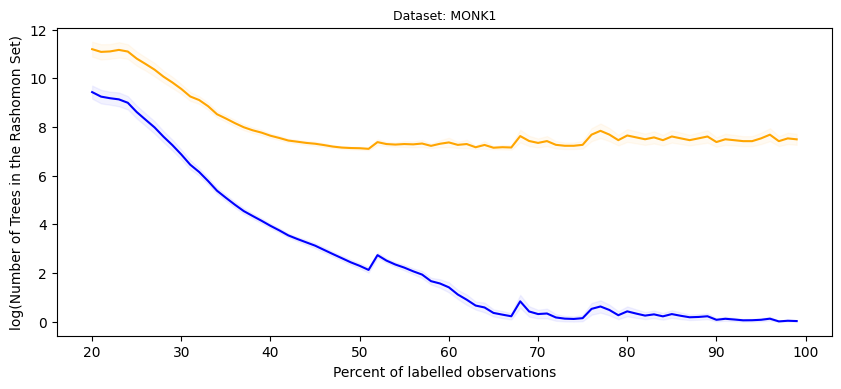}
        \end{subfigure}
        \vspace{1em}
        %%% Row 3: MONK3  %%%
        \begin{subfigure}[t]{\columnwidth}
            \centering
            \includegraphics[width=\columnwidth]{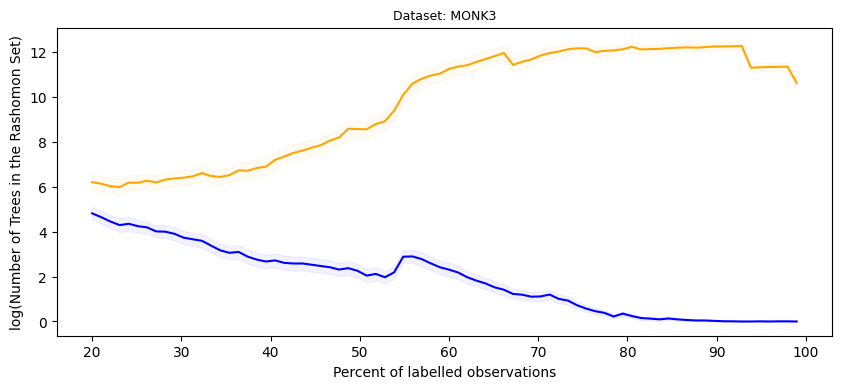}
        \end{subfigure}
    \end{minipage}
    \hfill
    %%% Right Column %%%
    \begin{minipage}[t]{0.45\textwidth}
        %%% Row 4: Bar7 %%%
        \begin{subfigure}[t]{\columnwidth}
            \centering
            \includegraphics[width=0.9\columnwidth]{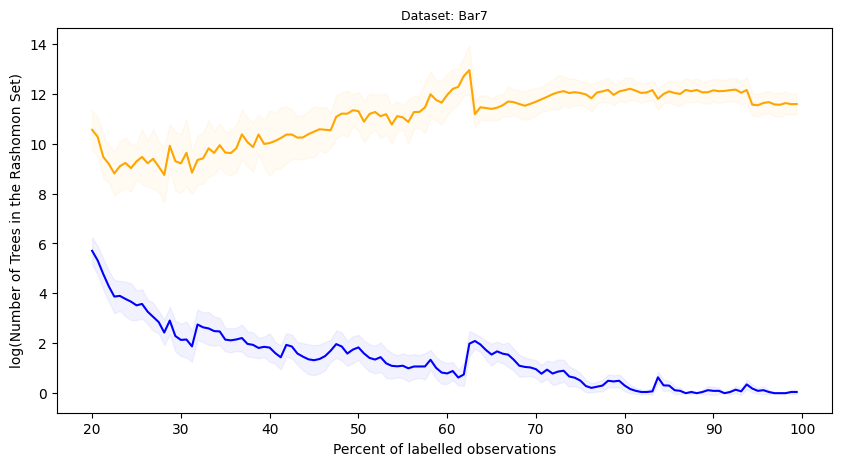}
        \end{subfigure}
        \vspace{1em}
        %%% Row 5: COMPAS  %%%
        \begin{subfigure}[t]{\columnwidth}
            \centering
            \includegraphics[width=0.9\columnwidth]{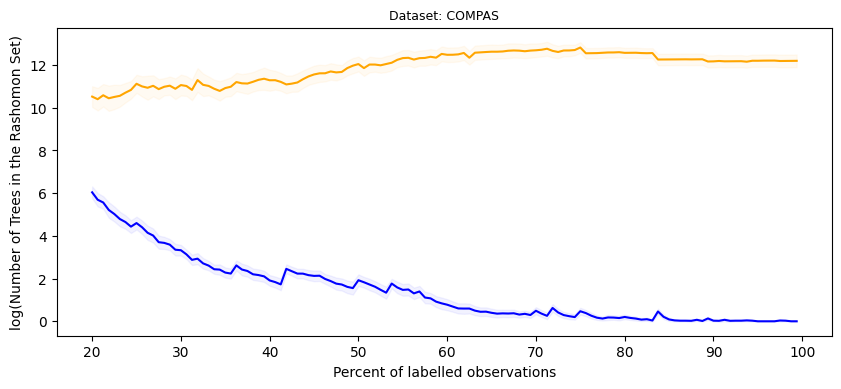}
        \end{subfigure}
        \vspace{3em} % Extra space to align with left column's 3 figures
    \end{minipage}
    
    %%% Legend below all plots %%%
    \begin{tikzpicture}
        % Define custom colors using RGB from Python
        \definecolor{PassiveColor}{rgb}{0, 0, 0}  % Black
        \definecolor{RFColor}{rgb}{0, 0.5, 0}     % Adjusted Green
        \definecolor{DUREALColor}{rgb}{1, 0.5, 0} % Orange
        \definecolor{UNREALColor}{rgb}{0, 0, 1}   % Blue
        
        % Adjusted positions for even spacing across figure width
        % \draw[PassiveColor, thick] (0,0) -- (0.5,0) node[right, black] {Passive};
        % \draw[RFColor, thick] (2.0,0) -- (2.5,0) node[right, RFColor] {RF};
        \draw[DUREALColor, thick] (1.0,0) -- (1.5,0) node[right, DUREALColor] {All Trees};
        \draw[UNREALColor, thick] (4,0) -- (4.5,0) node[right, UNREALColor] {Unique Trees};
    \end{tikzpicture}
    
    %%% Caption and Label %%%
    \caption{The number of total vs. unique trees on a logarithmic scale.}
    \label{fig:TreeGrid}
\end{figure}
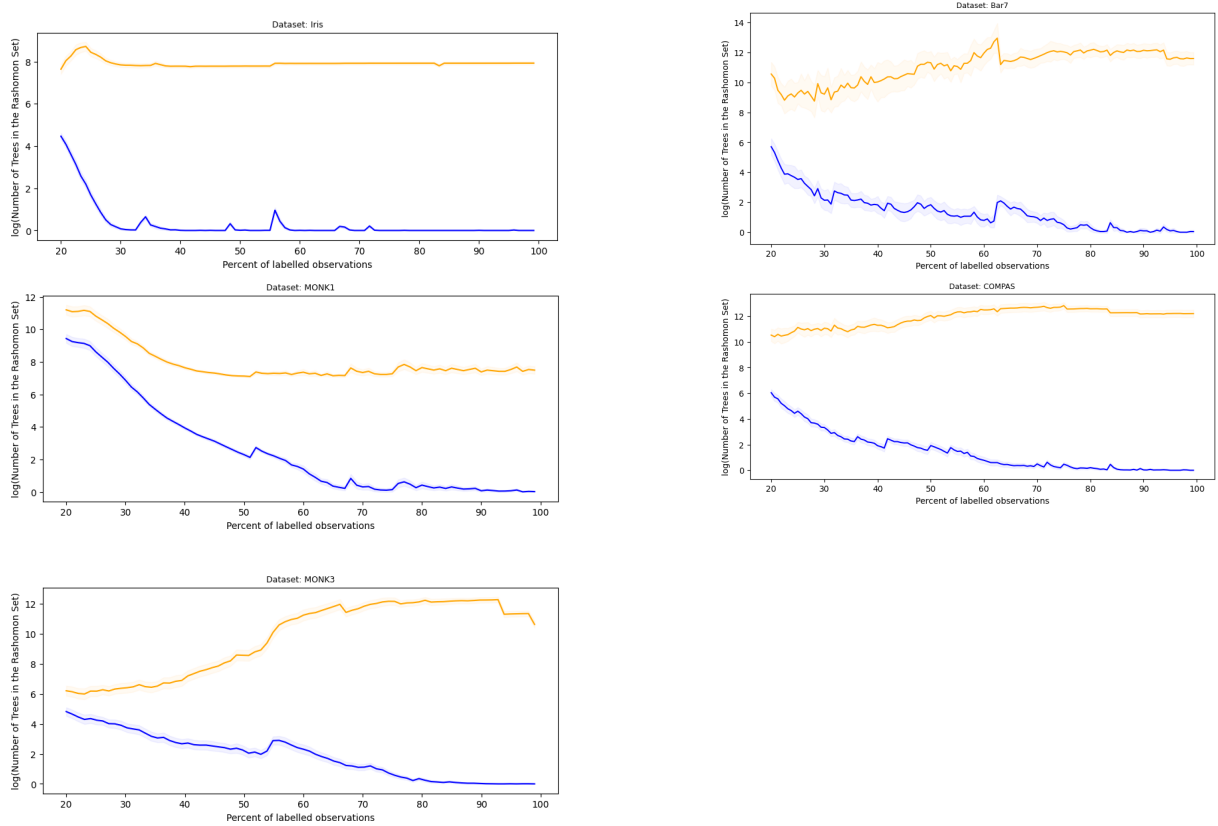

This relationship between tree count and unique classification patterns reveals a deeper insight into model diversity. As shown in Figure \ref{fig:TreeGrid}, increasing the number of trees does not proportionally increase the number of unique classification patterns. Further analysis in Section \ref{sec:UniqueVsDuplicate} demonstrates that while raising the Rashomon threshold increases the total tree count in the Rashomon set, the number of unique patterns remains relatively stable. As more data is gathered, the number of unique classification patterns decreases towards a few explanations while the number of redundant trees usually continues to grow. The reduction to a single explanation highlights that our method accounts for model ambiguity by strategically prioritizing the querying of observations with the highest uncertainty across models.

Both \textit{UNREAL} and \textit{DUREAL} leverage this principle effectively, though in different ways: while \textit{DUREAL} benefits from the implicit weighting of complex patterns, \textit{UNREAL} achieves similar performance with a more concise set of explanations. This distinction highlights an important principle: trees themselves are not explanations, but rather manifestations of underlying classification patterns.

%%% FIVE BY ONE FIGURES %%%

\section{Limitations and Future Work}
Our method, while demonstrating these promising results, presents certain limitations that warrant discussion and possible future work. One constraint lies in the initial selection of the Rashomon threshold. This initial threshold, though designed to optimize selection criteria in the early stages, may not maintain its optimality as the process evolves. As evidenced in Figure \ref{fig:TreeGrid}, the number of unique classification patterns exhibits considerable variation throughout the active learning process.

This limitation can be addressed in future work by dynamically recalibrating $\epsilon$ at each iteration of the active learning procedure. That is, after each query, researchers can again set a sufficiently high $\epsilon$ and generate a new tradeoff plot to identify the best threshold. This iterative approach allows for the Rashomon threshold to adapt to the evolving dataset. However, this iterative recalibration introduces additional computational overhead, making it practical only when substantial computational resources are available.

This computational intensity represents another significant limitation of our approach. As detailed in Table \ref{tab:DatasetTable}, which presents the average run time for each active learning strategy, our Rashomon-based methods may not be optimal for researchers who prioritize computational efficiency over predictive accuracy when compared to traditional machine learning models.

These limitations notwithstanding, our findings plant the seeds for future active learning research that incorporates the Rashomon set of good and plausible models. In particular, \citet{AparaRPS}'s Rashomon Partition Sets (RPS) offer a promising framework for comprehensively enumerating the Rashomon set without an inherent geometric structure. Investigating the use of RPS in active learning may further deepen our understanding of Rashomon's benefits in both prediction and interpretability.

\section{Conclusion}\label{ConclusionSection}
Our work offers three key insights. Firstly, we demonstrate that ensembling over the Rashomon set of decision trees enhances the active learning process by a significant margin, up to $20\%$ in some datasets. Unlike traditional ensemble methods which aggregate over the entire space of models, potentially including models that are poor performing or implausible, the Rashomon set only contains models with high posterior probability. This focused approach ensures that our committee consists of only strong and plausible models, whose disagreements naturally occur in the regions that matter most for active learning. These targeted disagreements provide a more robust measure of uncertainty, leading to a more efficient query selection.

Secondly, our approach provides a novel mechanism for dealing with label noise in the active learning process. By leveraging the diversity of the Rashomon set, we can more effectively distinguish between genuine model uncertainty and noise-induced variability, thereby improving the robustness of the active learning strategy in challenging, noisy environments.

Finally, we address the issue of redundant and duplicate explanations when constructing a Rashomon set by only considering trees with unique explanations. Redundant explanations can inflate query-by-committee metrics and obscure interpretability. By only ensembling over the Rashomon's subset of trees with unique explanations, we ensure that the ensemble remains parsimonious and interpretable while maintaining high prediction accuracy.

% \clearpage

\bibliography{bibliography}
\bibliographystyle{icml2025}

%%%%%%%%%%%%%%%%%%%%%%%%%%%%%%%%%%%%%%%%%%%%%%%%%%%%%%%%%%%%%%%%%%%%%%%%%%%%%%%
%%%%%%%%%%%%%%%%%%%%%%%%%%%%%%%%%%%%%%%%%%%%%%%%%%%%%%%%%%%%%%%%%%%%%%%%%%%%%%%
% APPENDIX
%%%%%%%%%%%%%%%%%%%%%%%%%%%%%%%%%%%%%%%%%%%%%%%%%%%%%%%%%%%%%%%%%%%%%%%%%%%%%%%
%%%%%%%%%%%%%%%%%%%%%%%%%%%%%%%%%%%%%%%%%%%%%%%%%%%%%%%%%%%%%%%%%%%%%%%%%%%%%%%
\newpage
\appendix
\onecolumn

\section{Appendix}

\subsection{Geometry of trees in Group 1}
\begin{figure*}[!htb]
    \centering
    \begin{subfigure}[b]{0.25\textwidth}
        \centering
        \includegraphics[width=0.9\textwidth]{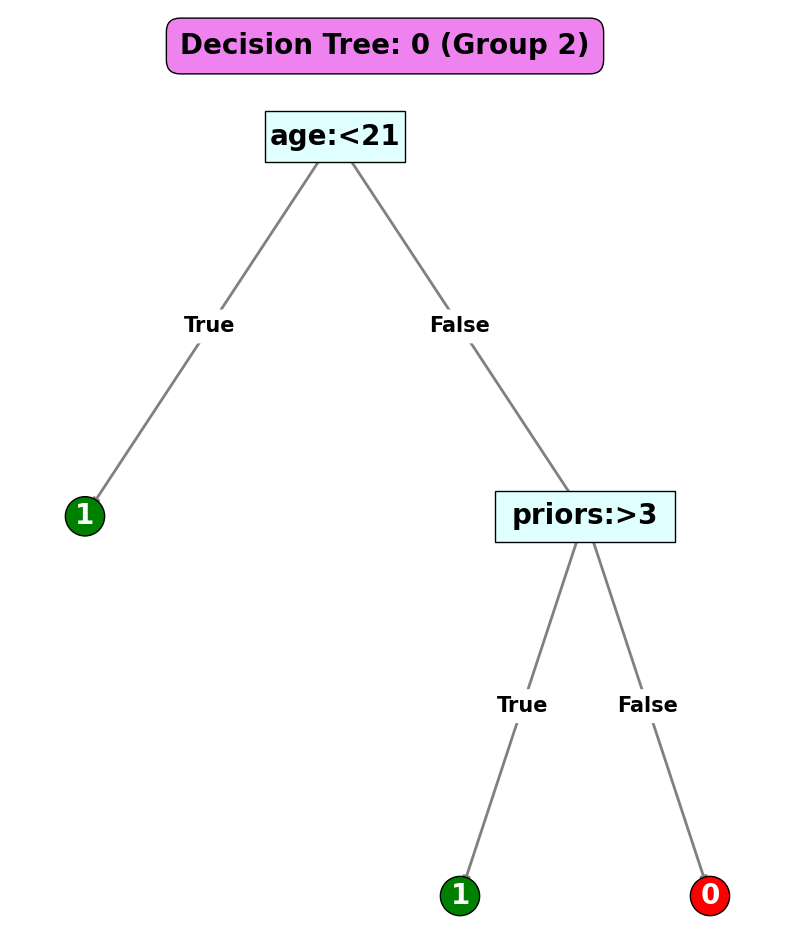}
    \end{subfigure}
    \hfill
    \begin{subfigure}[b]{0.25\textwidth}  
        \centering 
        \includegraphics[width=0.9\textwidth]{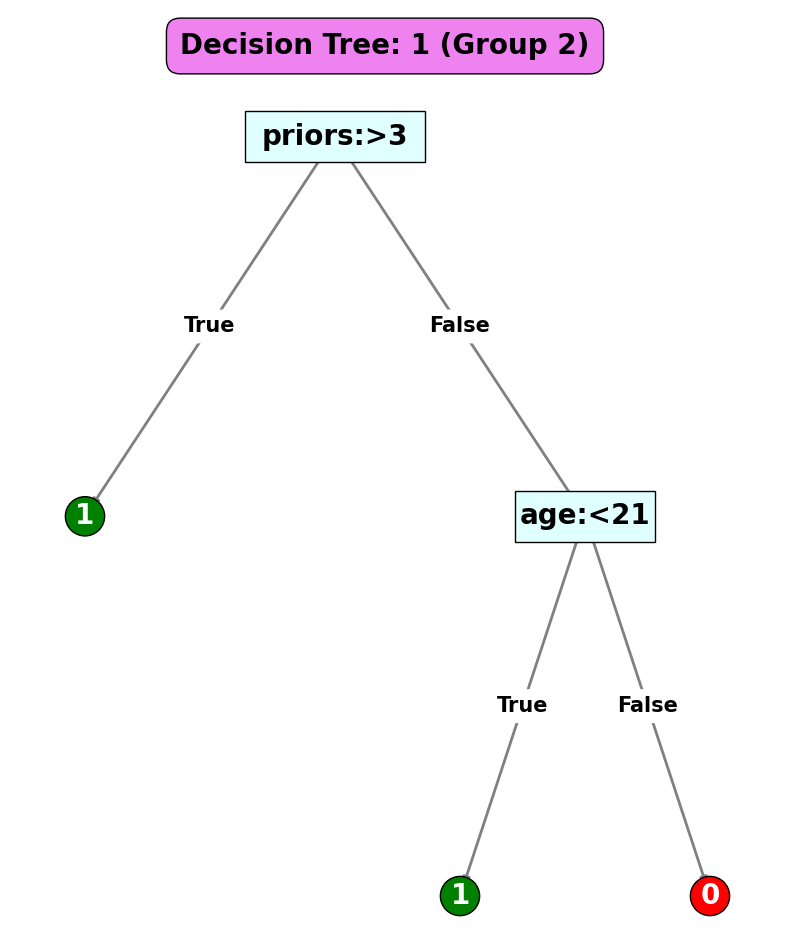}
    \end{subfigure}
    \hfill
    \begin{subfigure}[b]{0.25\textwidth}   
        \centering 
        \includegraphics[width=0.9\textwidth]{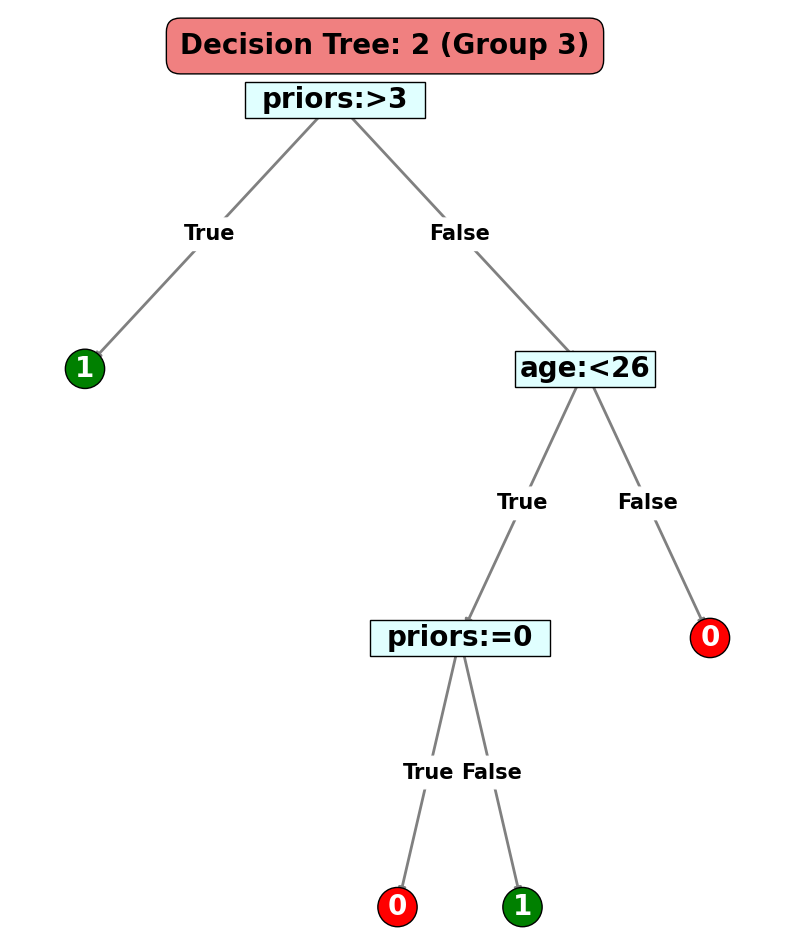}
    \end{subfigure}
    \vskip\baselineskip
    \begin{subfigure}[b]{0.25\textwidth}   
        \centering 
        \includegraphics[width=0.9\textwidth]{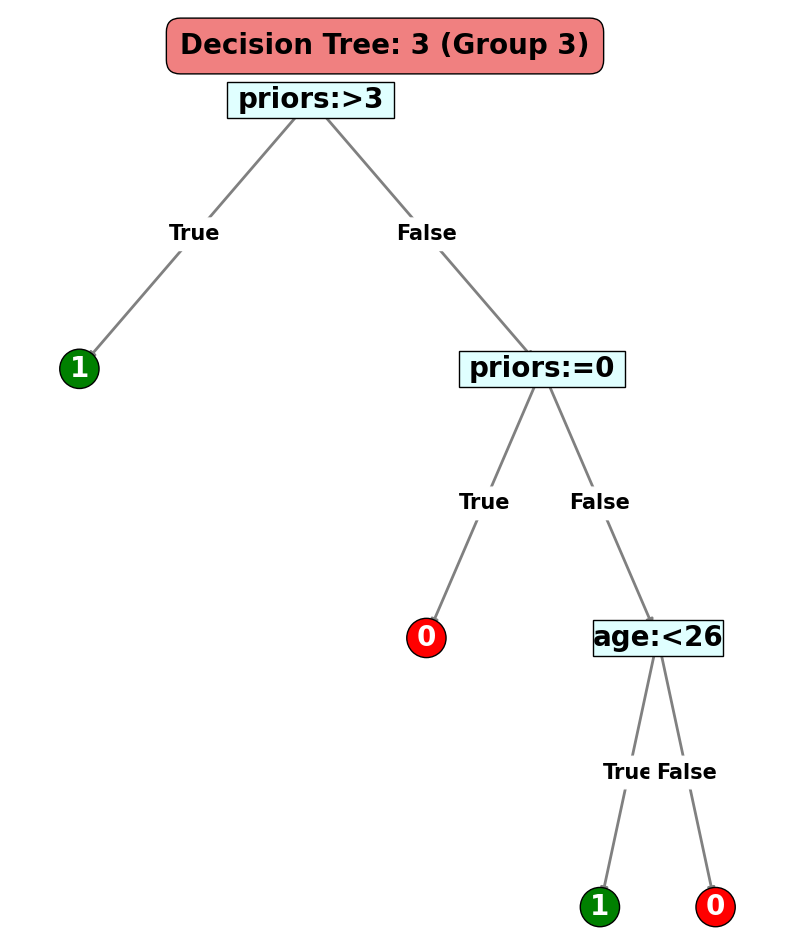}
    \end{subfigure}
    \hfill
    \begin{subfigure}[b]{0.25\textwidth}   
        \centering 
        \includegraphics[width=0.9\textwidth]{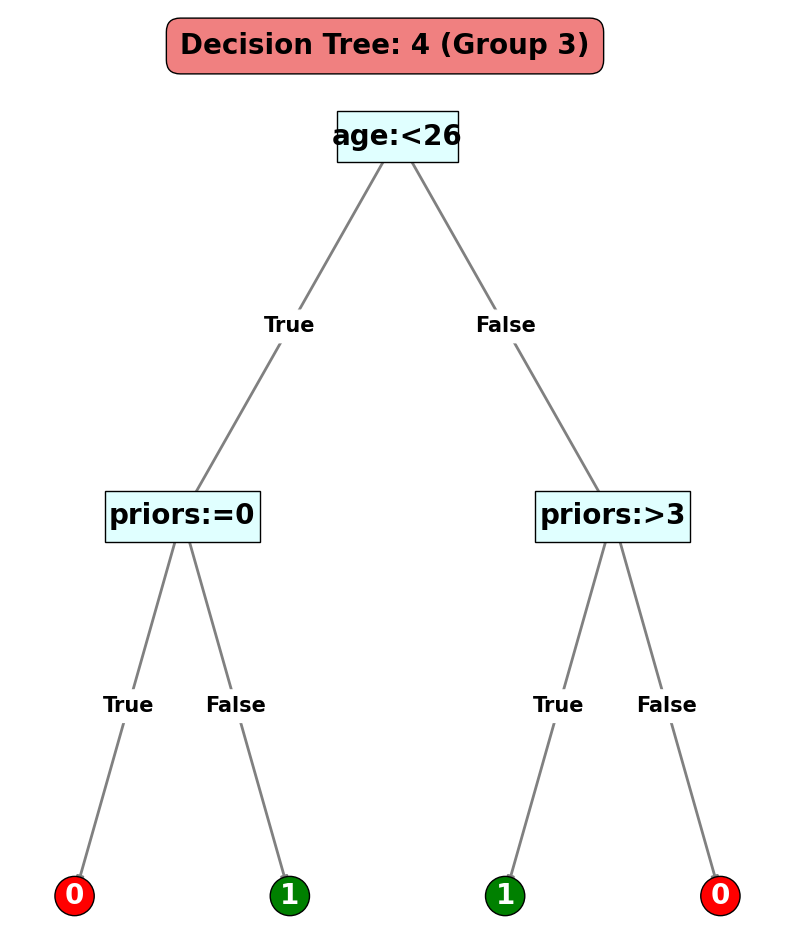}
    \end{subfigure}
    \hfill
    \begin{subfigure}[b]{0.25\textwidth}   
        \centering 
        \includegraphics[width=0.9\textwidth]{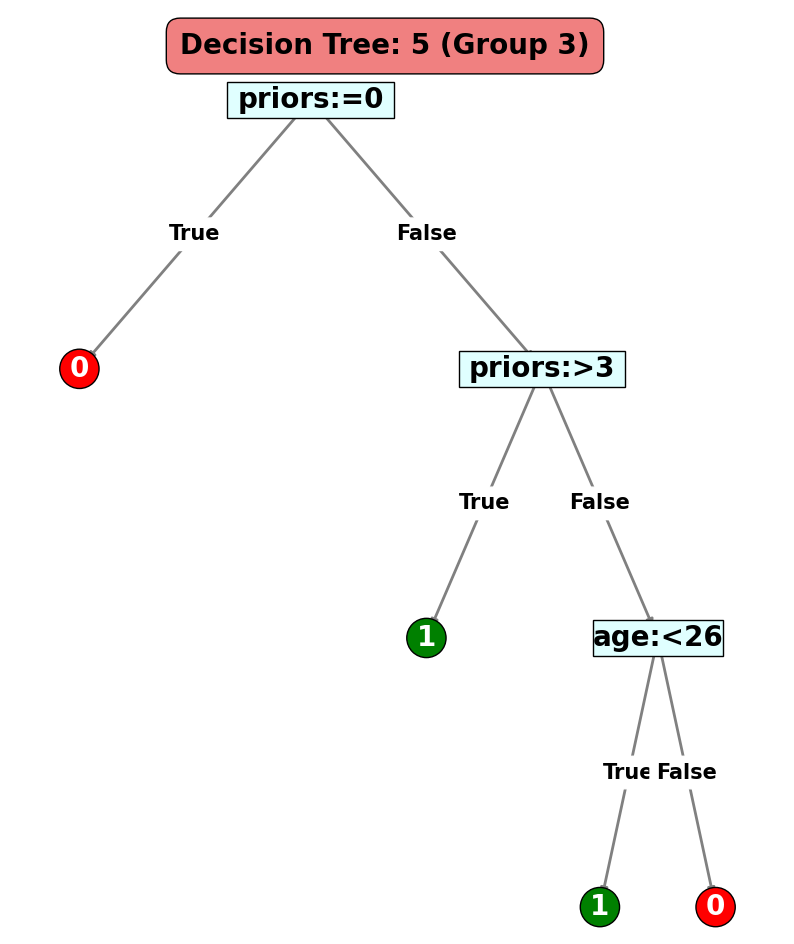}
    \end{subfigure}
    \vskip\baselineskip
    \begin{subfigure}[b]{0.25\textwidth}   
        \centering 
        \includegraphics[width=0.9\textwidth]{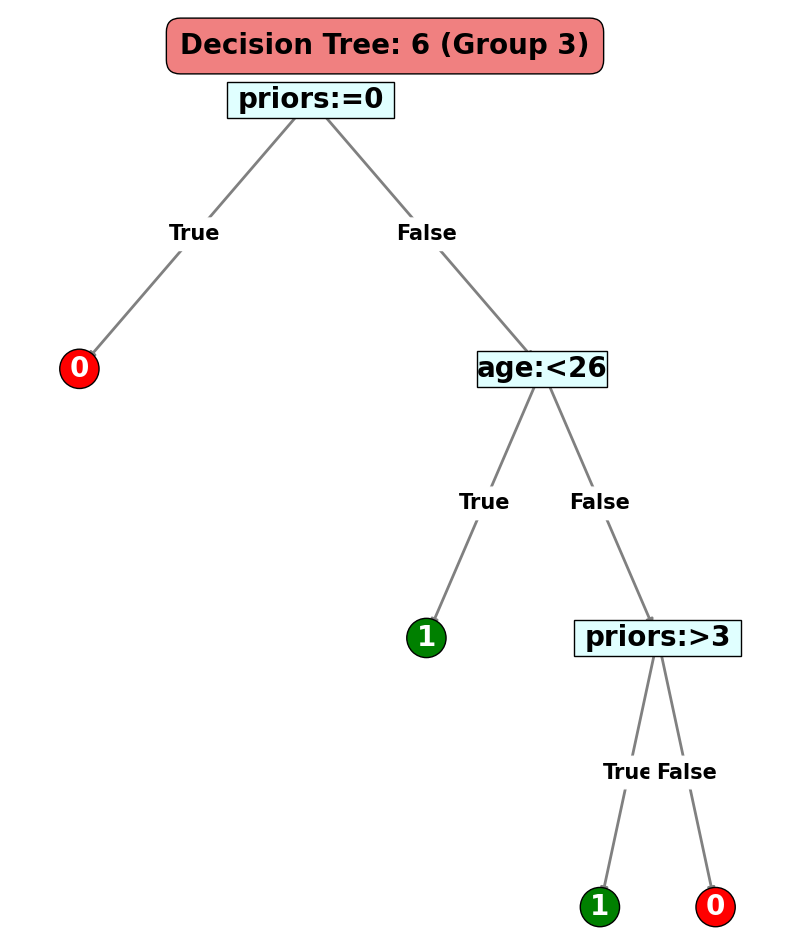}
    \end{subfigure}
    \hfill
    \begin{subfigure}[b]{0.25\textwidth}   
        \centering 
        \includegraphics[width=0.9\textwidth]{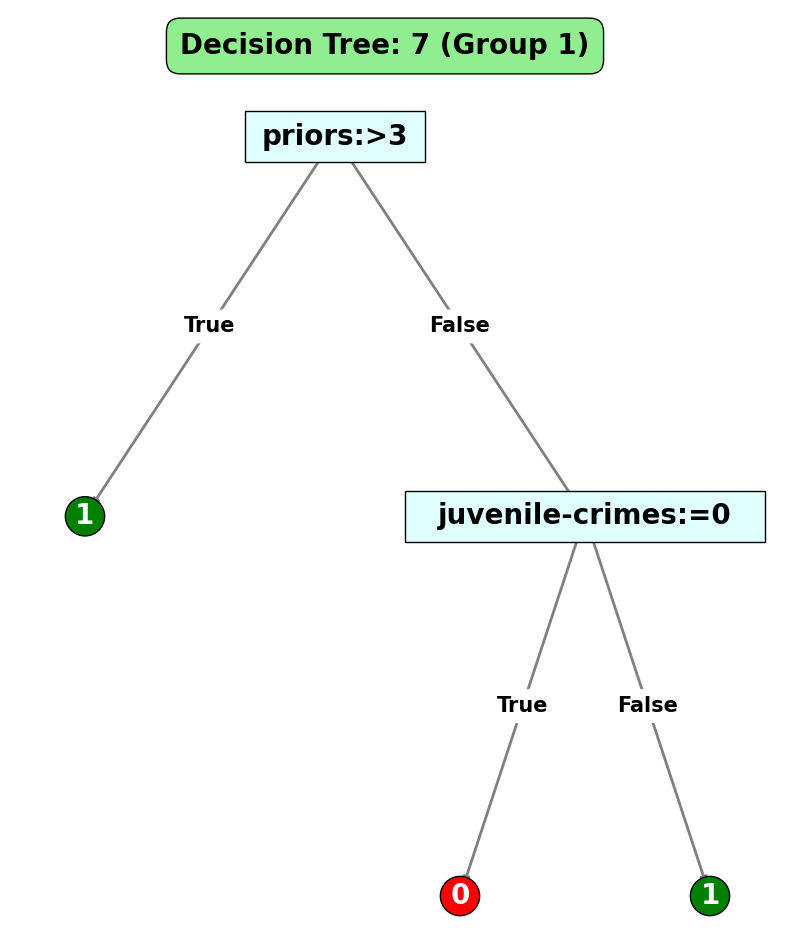}
    \end{subfigure}
    \hfill
    \begin{subfigure}[b]{0.25\textwidth}   
        \centering 
        \includegraphics[width=0.9\textwidth]{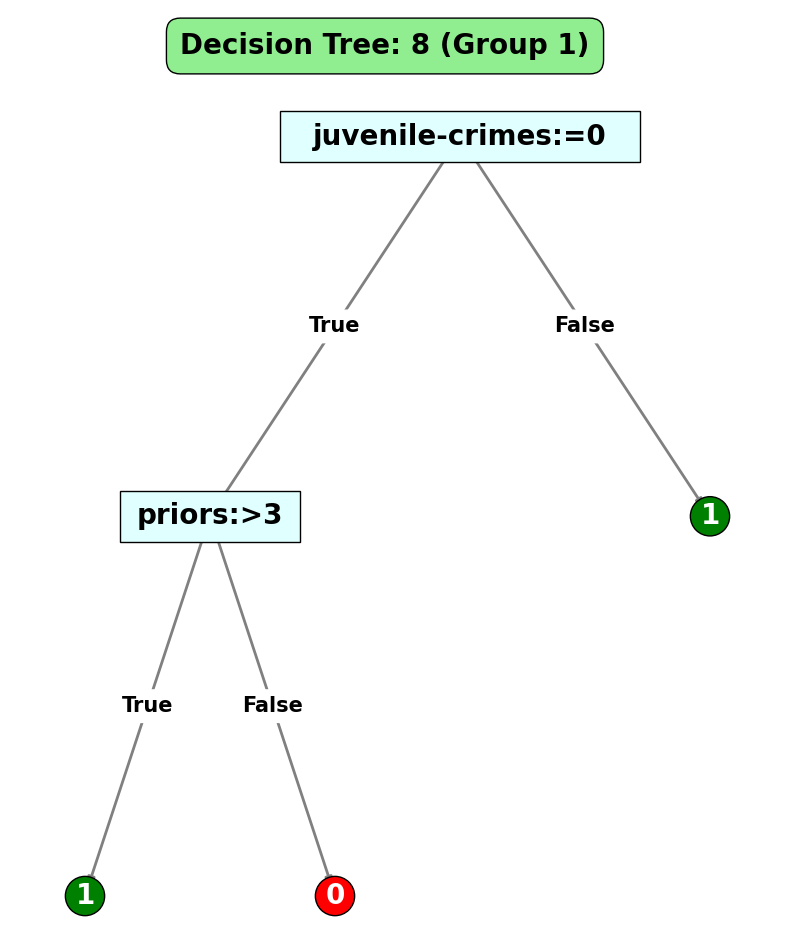}
    \end{subfigure}
    \caption{Geometry of the decision trees from Figure \ref{fig:ErrorbyTreeIndex_Grouped}. Note that these near-optimal trees, although sharing similar similar accuracy, differ in specific regions of the feature space, namely how to split the feature prior. In the \textit{UNREAL} algorithm, these trees would lead to a total of 4 members in the committee consisting of the four unique classification patterns. In \textit{DUREAL}, the committee would be composed of all nine trees.}
    \label{fig:TreeGeometry}
\end{figure*}

The images in Figure \ref{fig:TreeGeometry} give insight into the decision rules of the top 12 decision trees of Figure \ref{fig:ErrorbyTreeIndex_Grouped}. As seen, the trees exhibit very similar decision paths to each other, resulting in each one having the \textit{exact} same misclassification. As described in the main corpus, ensembling these trees as a committee and calculating the vote entropy metric off this committee will result in an inflated agreement and will recommend the observation that the best decision tree is most uncertain of rather than consider the uncertainty of the ensemble as a whole.

Note that Figure \ref{fig:ErrorbyTreeIndex_Grouped} categorizes trees by misclassification rate purely for visualization purposes. As detailed in Section \ref{Sec:GroupingTrees}, categorizing trees by misclassification error can clump together different distinct classification patterns that happen to share the same misclassification rate. It is important to note, then, to group and ensemble trees based on their distinct classification patterns in the \textit{UNREAL} algorithm. 

\subsection{Simulation Details}\label{SimulationDetailsAppendixSection}
\subsubsection{Data Preprocessing}
Table \ref{tab:DatasetTable} contains information on the datasets used in our simulations. 

We follow the same preprocessing of \citet{ExploringRashomonTrees} by one-hot encoding the numerical features whose thresholds were chosen by a gradient-boosted tree. The processed datasets were extracted directly from their \href{https://github.com/ubc-systopia/treeFarms/tree/main/experiments/datasets}{Github}. We also describe their preprocessing here. 

The MONK-1 and MONK-2 dataset were straightforward to one-hot encode. For the Iris dataset, the function \texttt{qcut} from Python's \texttt{Pandas} package was used to cut the numerical features of Iris into three equal bins. 

% To get the best splits and thresholds for the Wisconsin Breast Cancer dataset, the features and thresholds were selected by the ones used in a gradient-boosted tree with 40 decision stumps. This yielded the following: this turned out to be ``Clump\textunderscore Thickness = 10”, ``Uniformity\textunderscore Cell\textunderscore Size=1”, ``Uniformity\textunderscore Cell\textunderscore Size=10”, ``Uniformity\textunderscore Cell\textunderscore Shape=1”, ``Marginal\textunderscore Adhesion=1”, ``Single\textunderscore Epithelial\textunderscore Cell\textunderscore Size=2”, ``Bare\textunderscore Nuclei=1”, ``Bare\textunderscore Nuclei=10”, ``Normal\textunderscore Nucleoli=1”, and ``Normal\textunderscore Nucleoli=10”.

The COMPAS dataset, abbreviated for Correctional Offender Management Profiling for Alternative Sanctions, is a dataset for predicting which individuals will be arrested within two years of prison release. The dataset was discretized into binary features in the same way as \citet{NEURIPS2019_ac52c626}. The binary variables are ``sex = Female”, ``age $<$ 21”, ``age $<$ 23”, age $<$ 26, ``age $<$ 46”, ``juvenile felonies = 0”,
``juvenile misdemeanors = 0”, ``juvenile crimes = 0”, ``priors = 0”, ``priors = 1”, and ``priors = 2 to 3”, ``priors $>$ 3”.

The Bar7 dataset contains information on whether a customer will accept a coupon for a bar considering demographic and contextual attributes. The dataset was discretized as follows: ``Bar = 1 to 3”, ``Bar = 4 to 8”, ``Bar = less1”,
``maritalStatus = Single”, ``childrenNumber = 0”, ``Bar = gt8”, ``passenger = Friend(s)”,
``time = 6PM”, ``passenger = Kid(s)”, ``CarryAway = 4 to 8”, ``gender = Female”, ``education
= Graduate degree (Masters Doctorate etc.)”, ``Restaurant20To50 = 4 to 8”, ``expiration =
1d”, and ``temperature = 55”.

The Bar7 and COMPAS datasets were reduced to a fixed 200 observations due to computation time.

\begin{table}[t]
    \caption{Summary of our datasets.}
    \label{tab:DatasetTable}
    \vskip 0.15in
    \begin{center}
    \begin{small}
    \begin{sc}
    \begin{tabular}{l |c c c c c}
        \toprule
        & Iris & MONK1 & MONK3 & Bar7 & COMPAS \\
        \midrule
        Binary Features & 15 & 11 & 11 & 14 & 12 \\
        Test Set & 30 & 25 & 25 & 40 & 40 \\
        Initial Training Set & 24 & 19 & 19 & 32 & 32 \\
        Initial Candidate Set & 96 & 80 & 78 & 128 & 128 \\
        Rashomon Threshold $\epsilon$ & 0.025 & 0.030 & 0.019 & 0.020 & 0.025 \\
        UNREAL Mean Runtime (min.) & 3.26 & 39.27 & 131.86 & 340.66 & 512.35 \\
        DUREAL Mean Runtime (min.) & 3.23 & 38.63 & 132.66 & 350.18 & 546.79 \\
        \bottomrule
    \end{tabular}
    \end{sc}
    \end{small}
    \end{center}
    \vskip -0.1in
\end{table}

\subsubsection{Grouping unique classification patterns}\label{Sec:GroupingTrees}
Our \textit{UNREAL} algorithm depends on ensembling only the unique classification patterns from \texttt{TreeFarms}. This requires grouping trees into categories with identical classification patterns. To determine which trees should be grouped into similar classification patterns, we found the prediction of each tree for all observations in the candidate set. We then grouped trees together if each and every single prediction was the same. Note that grouping trees by their misclassification rate (as depicted in Figure \ref{alg:UNREAL} only for visualization purposes) may merge classification patterns.

% For instance, consider a triplet (Cell\textunderscore Size=1, Cell\textunderscore Size=10, Bare\textunderscore Nuclear=1) = (0,0,0). Placing any of the eight combinations through the trees of Figure \ref{fig:TreeGeometry} will result in a $1$ label in any of the trees of Group1A. Similarly, for triplet $(1,0,0)$ will will result in a $0$ for all trees in Group1A. This is true for any of the $8$ combinations. 

\subsubsection{Active learning simulation plots}
\begin{figure}
    \centering
    %%% Row 1: Iris  %%%
    \begin{subfigure}[t]{0.48\textwidth} % First column
        \centering
        \includegraphics[width=\textwidth]{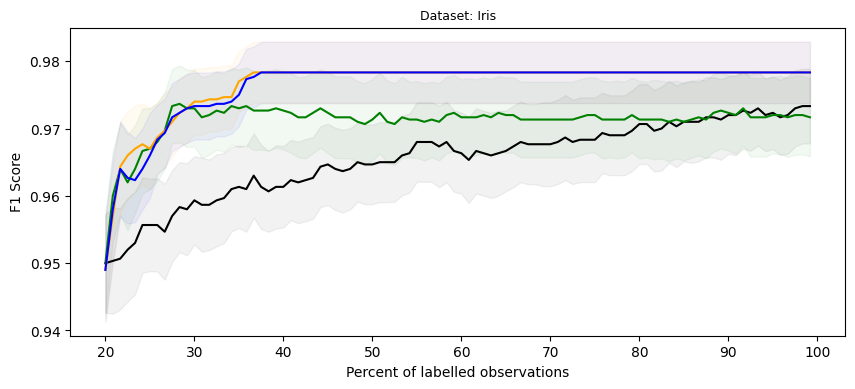}
    \end{subfigure}
    \hfill
    \begin{subfigure}[t]{0.48\textwidth} % First column
        \centering
        \includegraphics[width=\textwidth]{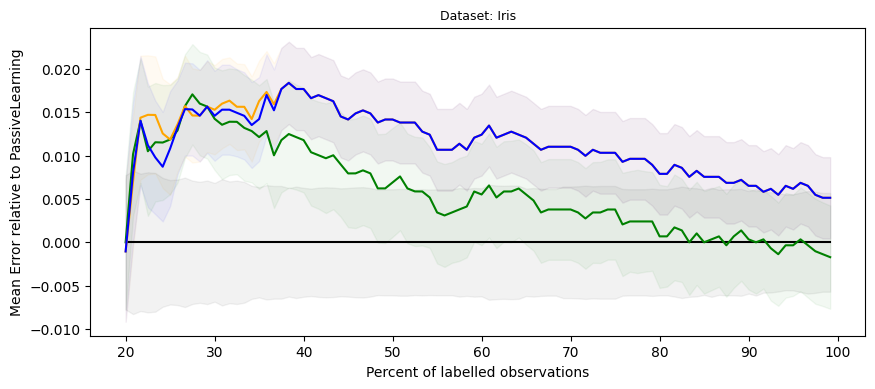}
    \end{subfigure}
    \vspace{1em}

    %%% Row 2: MONK1  %%%
    \begin{subfigure}[t]{0.48\textwidth} % First column
        \centering
        \includegraphics[width=\textwidth]{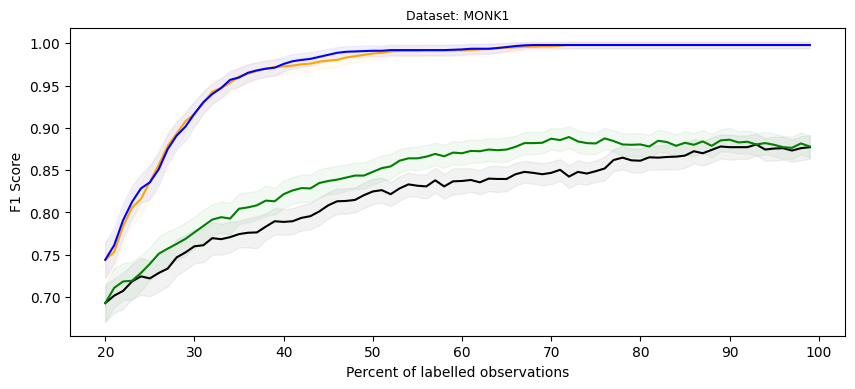}
    \end{subfigure}
    \hfill
    \begin{subfigure}[t]{0.48\textwidth} % First column
        \centering
        \includegraphics[width=\textwidth]{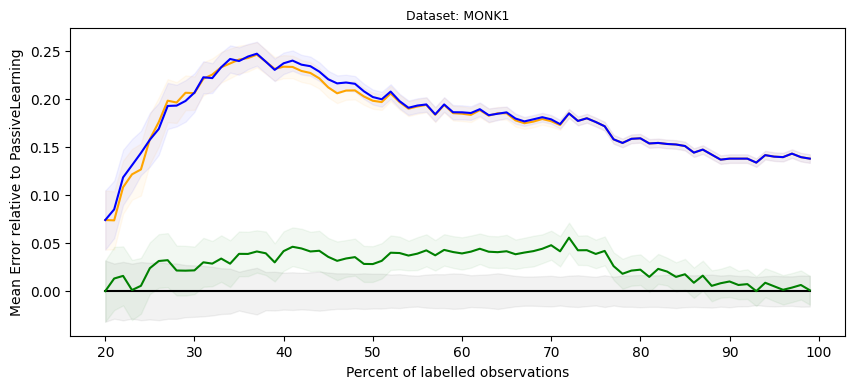}
    \end{subfigure}
    \vspace{1em}

    %%% Row 3: MONK3  %%%
    \begin{subfigure}[t]{0.48\textwidth} % First column
        \centering
        \includegraphics[width=\textwidth]{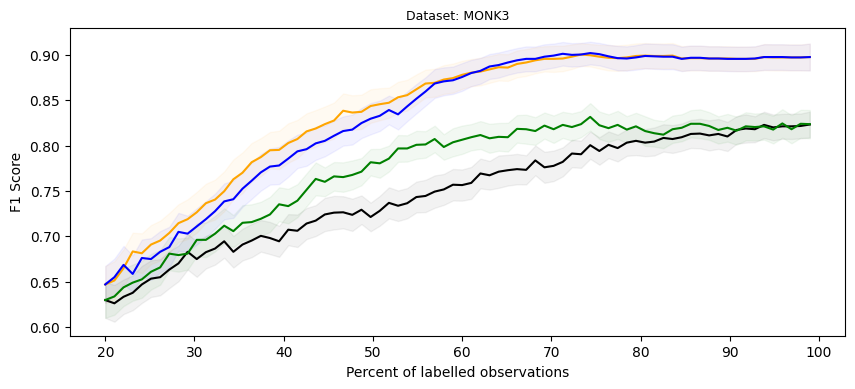}
    \end{subfigure}
    \hfill
    \begin{subfigure}[t]{0.48\textwidth} % First column
        \centering
        \includegraphics[width=\textwidth]{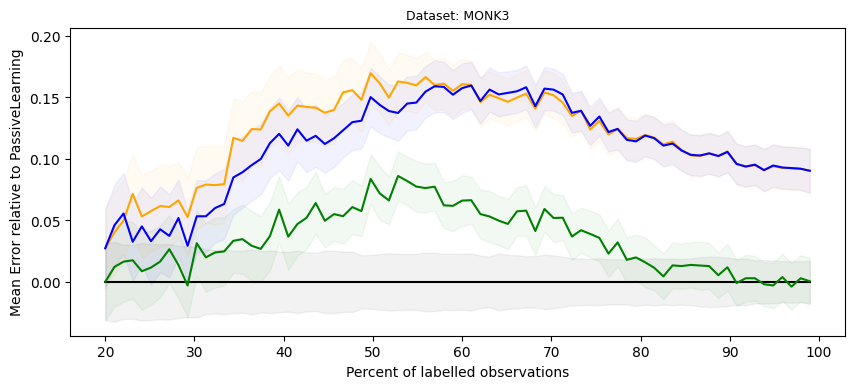}
    \end{subfigure}
    \vspace{1em}

    %%% Row 4: Bar7 %%%
    \begin{subfigure}[t]{0.48\textwidth} % First column
        \centering
        \includegraphics[width=\textwidth]{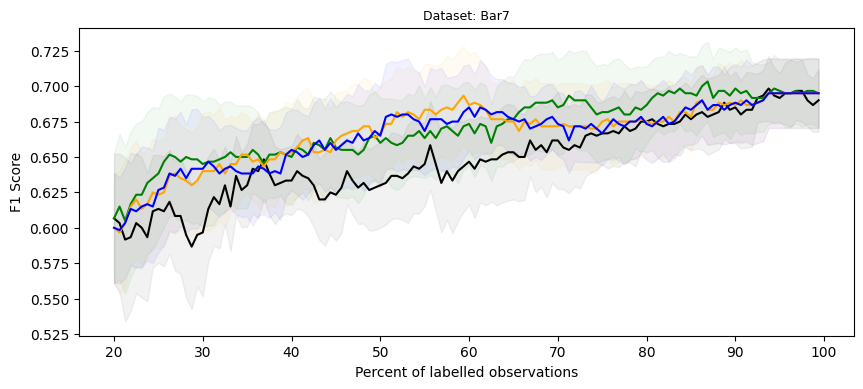}
    \end{subfigure}
    \hfill
    \begin{subfigure}[t]{0.48\textwidth} % First column
        \centering
        \includegraphics[width=\textwidth]{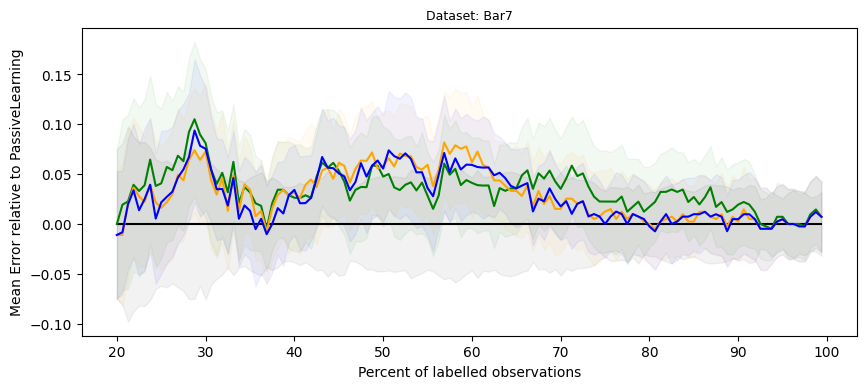}
    \end{subfigure}
    \vspace{1em}

    %%% Row 5: COMPAS  %%%
    \begin{subfigure}[t]{0.48\textwidth} % First column
        \centering
        \includegraphics[width=\textwidth]{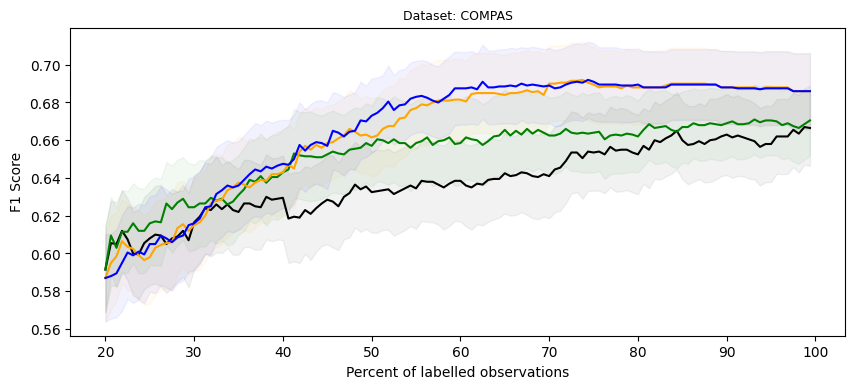}
    \end{subfigure}
    \hfill
    \begin{subfigure}[t]{0.48\textwidth} % First column
        \centering
        \includegraphics[width=\textwidth]{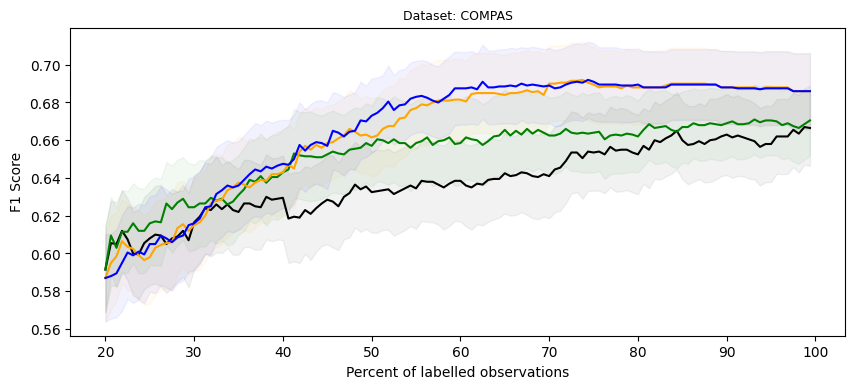}
    \end{subfigure}
    \vspace{1em}

    %%% Legend below all plots %%%
    \begin{tikzpicture}
        % Define custom colors using RGB from Python
        \definecolor{PassiveColor}{rgb}{0, 0, 0}  % Black
        \definecolor{RFColor}{rgb}{0, 0.5, 0}     % Adjusted Green
        \definecolor{DUREALColor}{rgb}{1, 0.5, 0} % Orange
        \definecolor{UNREALColor}{rgb}{0, 0, 1}   % Blue
        
        % Adjusted positions for even spacing across figure width
        \draw[PassiveColor, thick] (0,0) -- (0.5,0) node[right, black] {Passive};
        \draw[RFColor, thick] (2.0,0) -- (2.5,0) node[right, RFColor] {RF};
        \draw[DUREALColor, thick] (3.5,0) -- (4,0) node[right, DUREALColor] {DUREAL};
        \draw[UNREALColor, thick] (6.0,0) -- (6.5,0) node[right, UNREALColor] {UNREAL};
    \end{tikzpicture}
    
    %%% Caption and Label %%%
    \caption{Performance of the four active learning procedures on the five datasets without relative error (left) and relative to passive learning (right).}
    \label{fig:ActiveLearningErrorFigureRelativeToNone}
\end{figure}
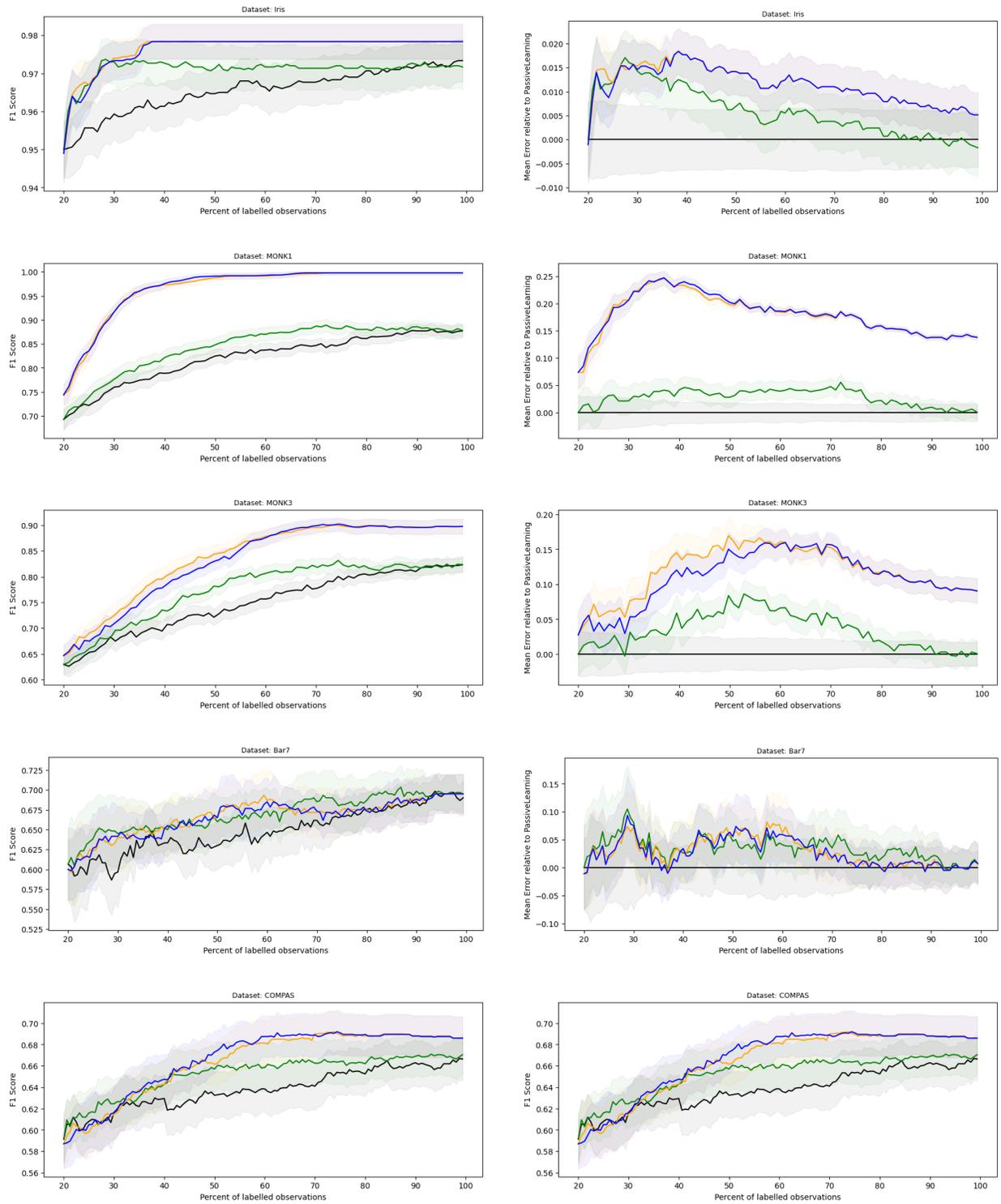

Additional active learning simulation plots without errors being relative to random forests and with errors relative to passive learning are presented in Figure \ref{fig:ActiveLearningErrorFigureRelativeToNone}.
%%% FIVE BY ONE FIGURES %%%

\subsubsection{Wilcoxon Rank Signed Test}\label{sec:WRST}
The following values present the p-values, rounded to 5 decimal points, of Wilcoxon ranked signed tests pairwise testing each active learning strategy.

\begin{table}[!htp]
    \centering
    \begin{tabular}{l llll}
        \toprule
        Dataset & PassiveLearning & RandomForest & UNREAL & DUREAL \\
        \midrule
        \multicolumn{5}{l}{\textbf{Iris}} \\
        PassiveLearning & 1.0 &  &  &  \\
        RandomForest & 0.0 & 1.0 &  &  \\
        UNREAL & 0.0 & 0.0 & 1.0 &  \\
        DUREAL & 0.0 & 0.0 & 0.00398 & 1.0 \\
        \midrule
        \multicolumn{5}{l}{\textbf{MONK-1}} \\
        PassiveLearning & 1.0 &  &  &  \\
        RandomForest & 0.0 & 1.0 &  &  \\
        UNREAL & 0.0 & 0.0 & 1.0 &  \\
        DUREAL & 0.0 & 0.0 & 6e-05 & 1.0 \\
        \midrule
        \multicolumn{5}{l}{\textbf{MONK-3}} \\
        PassiveLearning & 1.0 &  &  &  \\
        RandomForest & 0.0 & 1.0 &  &  \\
        UNREAL & 0.0 & 0.0 & 1.0 &  \\
        DUREAL & 0.0 & 0.0 & 2e-05 & 1.0 \\
        \midrule
        \multicolumn{5}{l}{\textbf{Bar7}} \\
        PassiveLearning & 1.0 &  &  &  \\
        RandomForest & 0.0 & 1.0 &  &  \\
        UNREAL & 0.0 & 1e-05 & 1.0 &  \\
        DUREAL & 0.0 & 0.00238 & 0.0073 & 1.0 \\
        \midrule
        \multicolumn{5}{l}{\textbf{COMPAS}} \\
        PassiveLearning & 1.0 &  &  &  \\
        RandomForest & 0.0 & 1.0 &  &  \\
        UNREAL & 0.0 & 0.0 & 1.0 &  \\
        DUREAL & 0.0 & 0.0 & 0.0 & 1.0 \\
        \bottomrule
    \end{tabular}
    \caption{Wilcoxon ranked signed test comparing the statistical differences between the errors of our active learning procedures rounded to 5 decimal points.}
    \label{tab:WRST_All}
\end{table}

\subsection{How the Rashomon Threshold Affects the Number of Trees and Classification Patterns}\label{sec:UniqueVsDuplicate}

\begin{figure*}[!htp]
    \centering

    %%% Row 1: Iris  %%%
    \begin{subfigure}[t]{0.48\textwidth} % First column
        \centering
        \includegraphics[width=\textwidth]{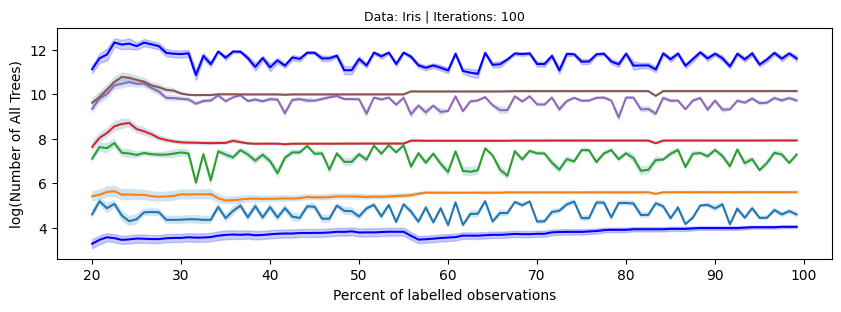}
    \end{subfigure}
    \hfill
    \begin{subfigure}[t]{0.48\textwidth} % Second column
        \centering
        \includegraphics[width=\textwidth]{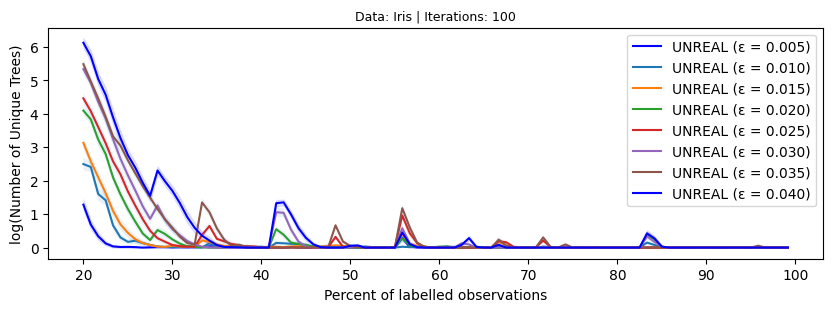}
    \end{subfigure}
    \vspace{1em}

    %%% Row 2: MONK1  %%%
    \begin{subfigure}[t]{0.48\textwidth} % First column
        \centering
        \includegraphics[width=\textwidth]{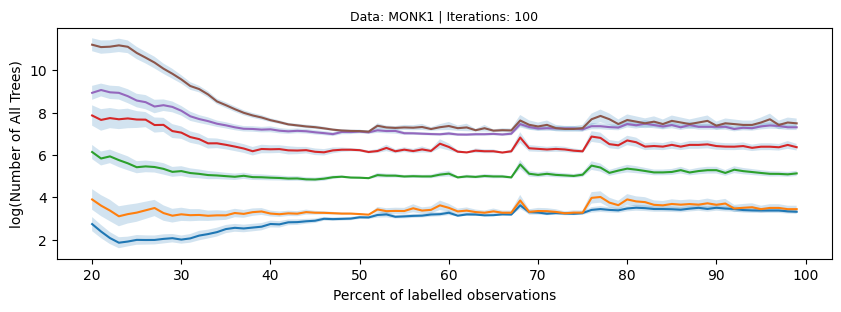}
    \end{subfigure}
    \hfill
    \begin{subfigure}[t]{0.48\textwidth} % Second column
        \centering
        \includegraphics[width=\textwidth]{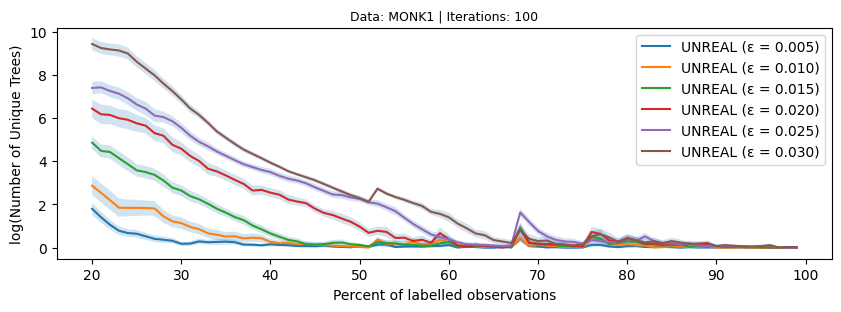}
    \end{subfigure}
    \vspace{1em}

    %%% Row 3: MONK3  %%%
    \begin{subfigure}[t]{0.48\textwidth} % First column
        \centering
        \includegraphics[width=\textwidth]{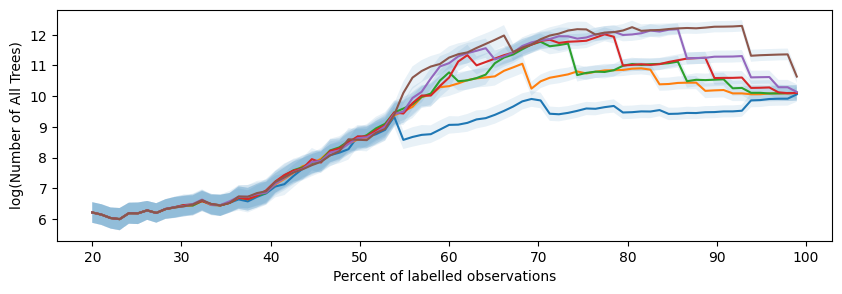}
    \end{subfigure}
    \hfill
    \begin{subfigure}[t]{0.48\textwidth} % Second column
        \centering
        \includegraphics[width=\textwidth]{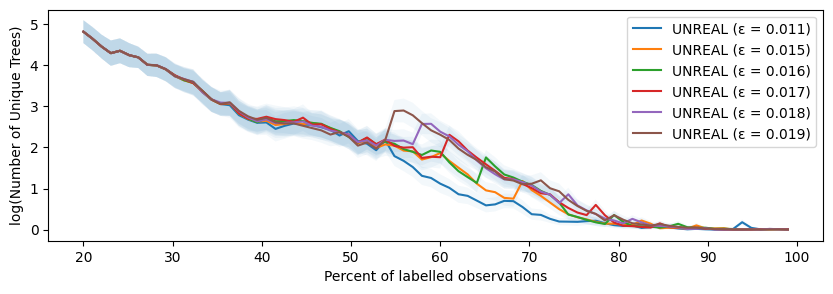}
    \end{subfigure}
    \vspace{1em}

    \caption{The left column indicates the log total number of Rashomon trees with the right indicating the log number of unique classification patterns.}
    \label{fig:UniqueDuplicateTreeCounts_Grid}
\end{figure*}

Increasing the Rashomon threshold will obviously increase the number of trees in the Rashomon set. However, we wanted to see if this would also increase the number of unique classification patterns in the Rashomon set. This exploration is important as diversifying our QBC committee with new and different unique classification patterns will affect the query-selection criteria. As such, at each iteration, we collect the total number of trees and the number of unique classification patterns. Figure \ref{fig:UniqueDuplicateTreeCounts_Grid} includes both of these for the following three datasets: Iris, MONK-1, and MONK-3. 
    
We find that increasing the Rashomon threshold greatly increases the number of trees in the Rashomon set, but not at the same magnitude as the number of classification patterns. The number of unique classification patterns is large at the beginning active learning process, but decreases as we begin to collect data. After collecting a sufficient amount of data points, the number of unique classification patterns is reduced to $1$. For instance, in the Iris dataset, unique classification patterns range between $e^{2}$ and $e^{10}$, but reduce down to $1$ when enough data is collected. The reduction to $1$ unique explanation highlights the fact that our method initially

\end{document}